%% file: main.tex
\documentclass[acmlarge,nonacm]{acmart}
\usepackage{subcaption}
\usepackage{wasysym}
\usepackage{makecell}
\usepackage{enumitem}
\usepackage{soul}
\usepackage{color, xcolor}
\usepackage{pifont}
\usepackage{placeins}
\usepackage[most]{tcolorbox}
\usepackage{colortbl} 

\definecolor{ao}{rgb}{0.0, 0.5, 0.0}
\definecolor{britishracinggreen}{rgb}{0.0, 0.26, 0.15}
\definecolor{carnelian}{rgb}{0.7, 0.11, 0.11}

\newcommand{\cmark}{{\color{ao} \ding{52}}}%
\newcommand{\xmark}{{\color{carnelian} \ding{56}}}%

\newcommand{\myparagraph}[1]{\vspace{1mm} \noindent \textbf{\textsf{#1}}}
\AtBeginDocument{%
  \providecommand\BibTeX{{%
    \normalfont B\kern-0.5em{\scshape i\kern-0.25em b}\kern-0.8em\TeX}}}

\setcopyright{none}

\newcommand{\cmt}[1]{\textcolor{black}{#1}}

\acmJournal{IMWUT}
\acmVolume{1}
\acmNumber{1}
\acmArticle{1}
\acmMonth{5}

\renewcommand{\authorsaddresses}{}
\renewcommand\footnotetextcopyrightpermission[1]{} 
\begin{document}

\newcommand{\workname}{ProAgent} 

\title{\workname: Harnessing On-Demand Sensory Contexts for Proactive Agent Systems in the Wild}

\makeatletter
\def\@fnsymbol#1{%
  \ifcase#1\relax 
    \or †
    \or ¶
    \or ‡
    \or §
    \or ¶
    \else
      \@arabic{#1}
  \fi}
\makeatother

\author{Bufang Yang}
\authornote{Co-primary authors.}
\affiliation{%
  \institution{The Chinese University of Hong Kong}
  \country{China}}
\email{bfyang@link.cuhk.edu.hk}

\author{Lilin Xu}
\authornotemark[1]
\affiliation{%
  \institution{Columbia University}
  \country{United States}}
\email{lx2331@columbia.edu}

\author{Liekang Zeng}
\affiliation{%
  \institution{The Chinese University of Hong Kong}
  \country{China}}
\email{lkzeng@cuhk.edu.hk}

\author{Yunqi Guo}
\affiliation{%
  \institution{The Chinese University of Hong Kong}
  \country{China}}
\email{yunqiguo@cuhk.edu.hk}

\author{Siyang Jiang}
\affiliation{%
  \institution{The Chinese University of Hong Kong}
  \country{China}}
\email{syjiang@ie.cuhk.edu.hk}

\author{Wenrui Lu}
\affiliation{%
  \institution{The Chinese University of Hong Kong}
  \country{China}}
\email{wenruilu@cuhk.edu.hk}

\author{Kaiwei Liu}
\affiliation{%
  \institution{The Chinese University of Hong Kong}
  \country{China}}
\email{1155189693@link.cuhk.edu.hk}

\author{Yixuan Li}
\affiliation{%
  \institution{The Chinese University of Hong Kong}
  \country{China}}
\email{1155191356@link.cuhk.edu.hk}

\author{Xiaofan Jiang}
\affiliation{%
  \institution{Columbia University}
  \country{United States}}
\email{jiang@ee.columbia.edu}

\author{Guoliang Xing}
\affiliation{%
  \institution{The Chinese University of Hong Kong}
  \country{China}}
\email{glxing@cuhk.edu.hk}

\author{Zhenyu Yan}
\affiliation{%
  \institution{The Chinese University of Hong Kong}
  \country{China}}
\email{zyyan@ie.cuhk.edu.hk}

\begin{abstract}
Recent studies have begun to explore proactive large language model (LLM) agents that provide unobtrusive assistance by automatically leveraging contextual information, such as in code editing and in-app suggestions. 
However, most focus on short, task-specific episodes or on-screen contexts, rather than continuously perceiving and assisting users throughout daily life. 
Enabling such in-the-wild assistance requires continuous sensing of users’ surroundings, which can incur substantial system overhead.
In this work, we propose \workname, an end-to-end proactive agent system that harnesses on-demand sensory contexts to provide in-the-wild assistance.
\workname~first employs on-demand tiered perception to continuously sense users’ surroundings by integrating low-cost contextual cues with richer perception on demand, and uses proactive-oriented context extraction to derive hierarchical contexts integrating both sensory contexts and human preferences.
\workname~then employs a context-aware proactive reasoner to infer user needs and invokes external tools to deliver proactive assistance.
We implement \workname~on AR glasses and evaluate it on a public dataset and a real-world dataset. 
Results demonstrate that \workname~achieves up to 27.7\% higher proactive prediction accuracy and 20.5\% lower false detection than state-of-the-art baselines. 
A user study with 20 participants shows that 85\% were satisfied with \workname~and willing to use it in daily life.

\end{abstract}

\settopmatter{printfolios=true}
\maketitle
\pagestyle{plain}

\input{secs/1_intro}
\input{secs/2_related_works}

\input{secs/3_motivation}
\input{secs/4_system}

\input{secs/5_evaluation}
\input{secs/6_discussion}

\section{Conclusion}
We introduce \workname, an end-to-end proactive agent system that enables continuous, in-the-wild assistance in everyday life by integrating massive sensory contexts from mobile and wearable devices with LLM reasoning
Extensive evaluations on public and real-world datasets show that \workname~significantly improves proactive assistance quality, while our user study further demonstrates its effectiveness and user acceptance for daily use. We implement \workname~on AR glasses, and our evaluations show that \workname~achieves up to 27.7\% higher proactive prediction accuracy than state-of-the-art baselines, with 85\% of user-study participants willing to use it.

\bibliographystyle{ACM-Reference-Format}
\bibliography{main}

\newpage
\appendix
\section{Appendix: Prompt Settings}
\label{appendix}
\FloatBarrier

\begin{figure}[!htbp]
  \centering
\includegraphics[width=0.85\linewidth]{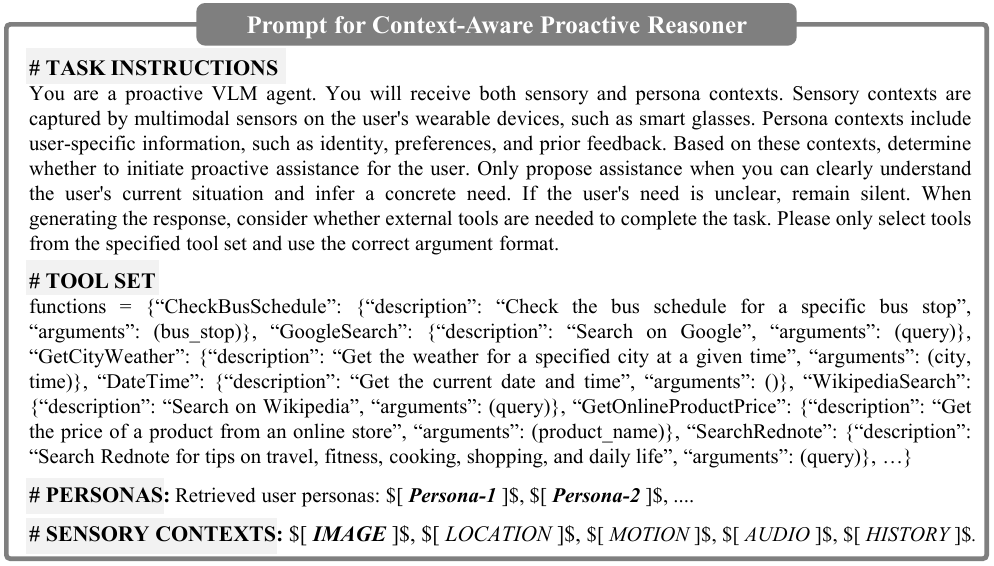}
\vspace{-1em}
  \caption{
Prompt for the context-aware proactive reasoner.}
\label{fig:prompt_reasoner}
\end{figure}

\begin{figure}[!htbp]
  \centering
\includegraphics[width=0.85\linewidth]{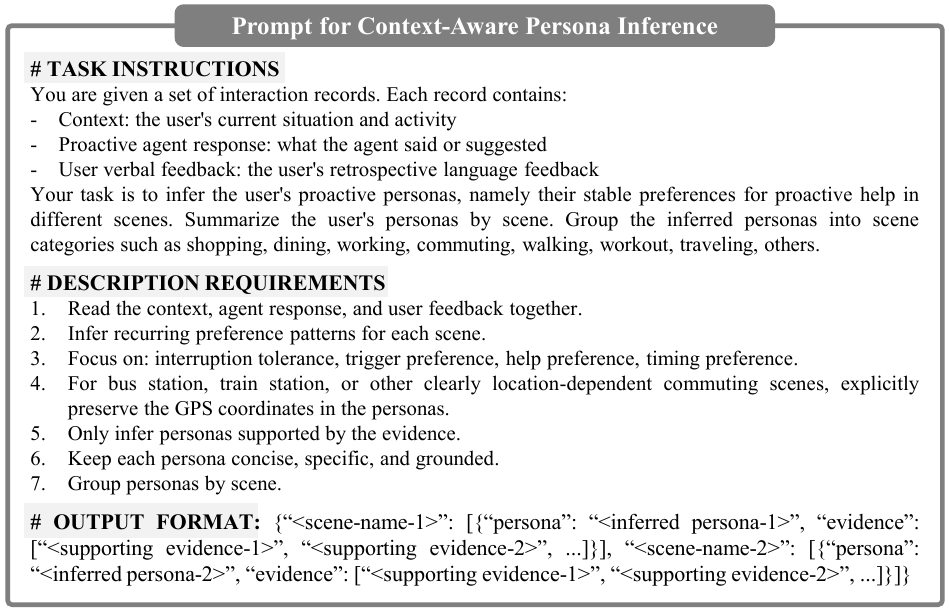}
\vspace{-1em}
  \caption{
Prompt for the context-aware persona inference.}
\label{fig:prompt_persona}
\end{figure}

\begin{figure}[!htbp]
  \centering
\includegraphics[width=0.85\linewidth]{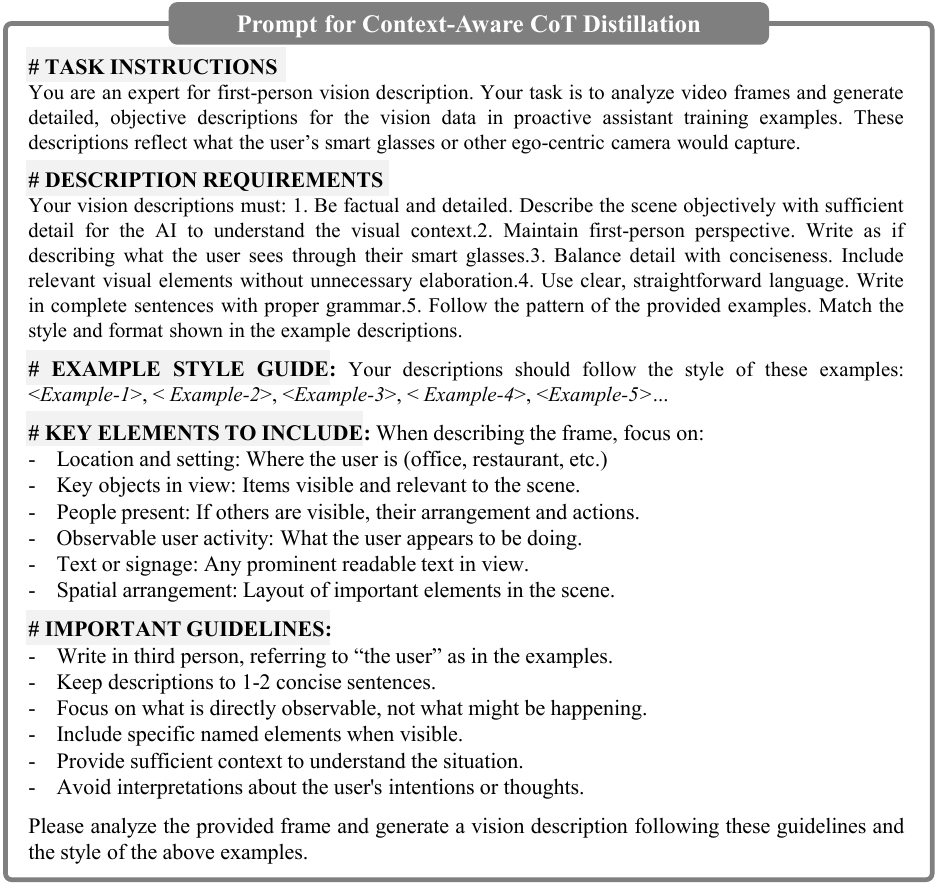}
\vspace{-1em}
  \caption{
Prompt for the context-aware CoT distillation.}
\label{fig:prompt_cot}
\end{figure}

\begin{figure}[!htbp]
  \centering
\includegraphics[width=0.85\linewidth]{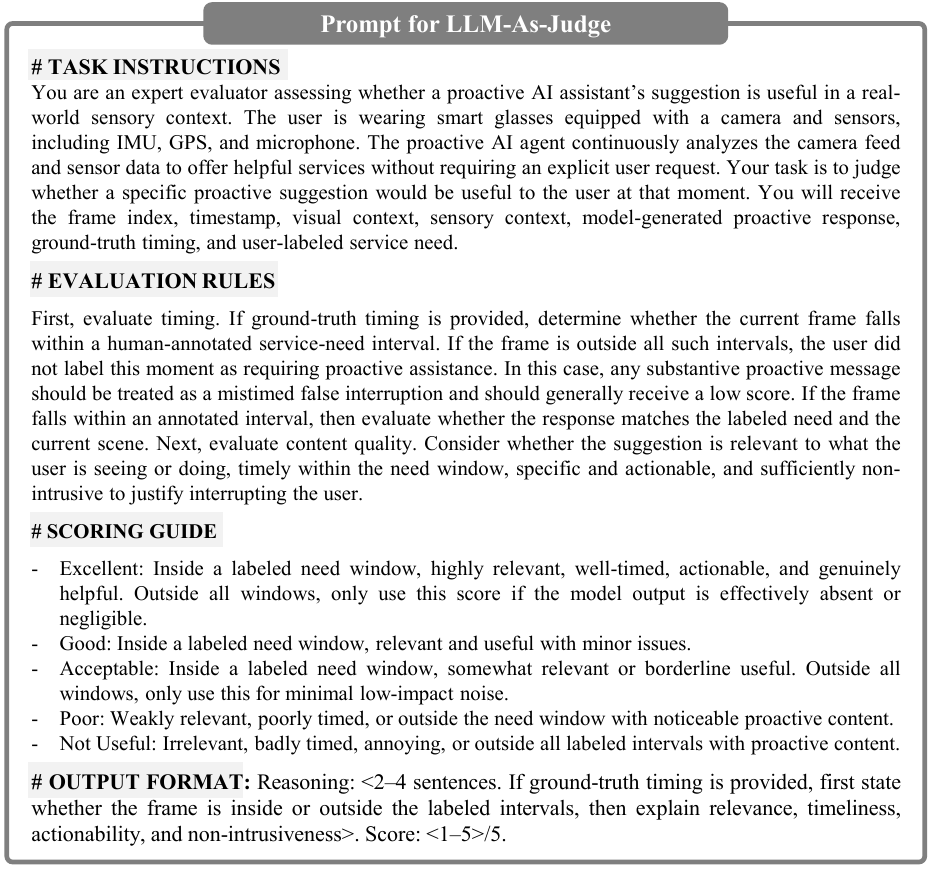}
\vspace{-1em}
  \caption{
Prompt for LLM-as-judge.}
\label{fig:prompt_eval}
\end{figure}

\end{document}

%% file: secs/1_intro.tex
\section{Introduction}
\begin{figure}
  \centering
\includegraphics[width=1\linewidth]{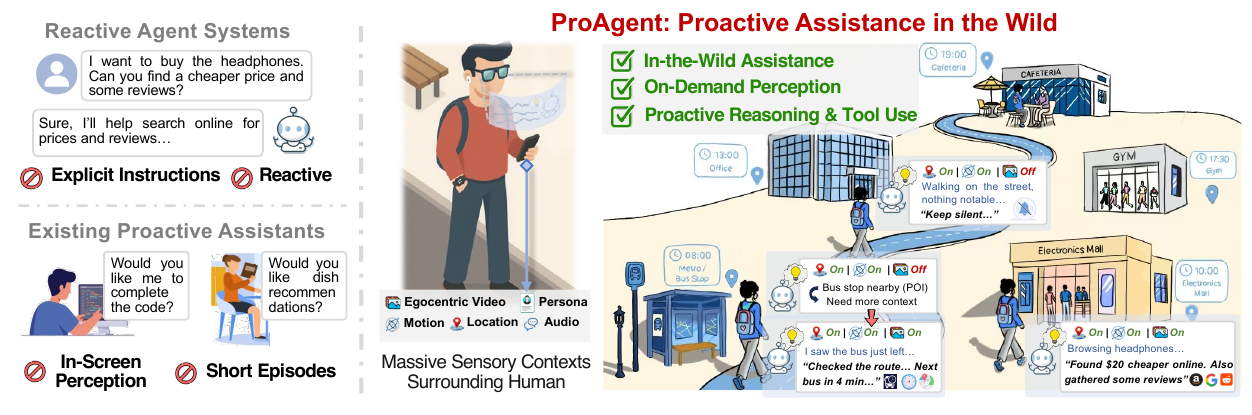}
\vspace{-2.5em}
  \caption{
  A user scenario of \workname. \workname~aims to enable longitudinal in-the-wild proactive assistance through on-demand perception, context-aware proactive reasoning, and tool calling, providing timely and unobtrusive support when appropriate.}
\label{fig:teaser}
  \vspace{-1.5em}
\end{figure}

Large Language Model (LLM) agents~\cite{li2024personal} have been increasingly explored as personal assistants, enabling applications such as coding assistants~\cite{yang2024embedgenius,shen2025autoiot}, smart-home control agents~\cite{king2024sasha,cheng2024autoiot}, and GUI agents~\cite{wen2024autodroid,zhang2023appagent}.
Recently, \textit{proactive agents} have emerged as a new class of LLM agents that anticipate user needs and initiate assistance without explicit requests~\cite{lu2024proactive,yang2025contextagent,zhao2025codinggenie}.
For example, proactive agents have been applied to code editors, desktop environments, and mobile interfaces, where they infer user intent and proactively provide timely suggestions or actions~\cite{lu2024proactive,magic-cue,zhao2025codinggenie}.
This shift from reactive agents toward proactive assistant systems that can perceive, reason, and assist users unobtrusively holds great potential for future personal assistance.

While early proactive assistants provide basic proactivity through predefined rules and conditions, such as context-aware reminders and health alerts~\cite{fall_detection_apple_watch,lei2025watchguardian,dey2000cybreminder}, they lack the flexible reasoning and action capabilities of LLM agents.
Recent proactive LLM agents extend proactive assistance with LLM-based reasoning, yet many remain confined to desktop screens, code editors, or application interfaces~\cite{lu2024proactive,magic-cue}, which limits the contexts available for inferring user needs beyond computer-use scenarios.
Recent advances in wearable sensing create new opportunities to move proactive assistance beyond screen-based contexts~\cite{xu2024can,yang2026sensorpersona}. Continuous multimodal sensing around the user allows proactive assistants to perceive the surrounding environment in situ and deliver timely, unobtrusive assistance in the wild, such as providing the next-bus arrival time when the user approaches a bus stop.
Some recent proactive assistants incorporate sensor contexts from mobile and wearable devices~\cite{yang2025socialmind,pu2025promemassist,li2025satori,lee2025sensible}, while they primarily focus on short task episodes and predefined scenarios, such as social interaction and cooking. 
This highlights a research gap in developing in-the-wild proactive LLM agents that can continuously leverage rich sensory contexts to understand users’ surroundings, infer their needs, and deliver timely, unobtrusive assistance.

To bridge this research gap, we aim to develop proactive LLM agents that can continuously perceive users’ surrounding sensory contexts while providing timely, unobtrusive assistance throughout daily life, as illustrated in Figure~\ref{fig:teaser}. However, developing such a system introduces several unique challenges.
\textit{First}, since users’ activities and environments evolve continuously, proactive agents require continuous perception of users’ surroundings throughout daily life. 
However, continuously capturing image data consumes unreasonably large power and computation, which is not possible on mobile and wearable devices.
Existing adaptive perception approaches are typically task-specific and rely on fixed sensing policies or heuristics, limiting their ability to support open-ended agent reasoning~\cite{neseem2020adasense,lu2011speakersense,paruchuri2025egotrigger}. Therefore, enabling proactive agents to dynamically determine when to acquire rich sensory contexts on demand remains challenging.
\textit{Second}, continuous perception in the wild produces massive, multimodal, and heterogeneous sensory streams.
Unlike reactive LLM agents that rely on explicit user requests~\cite{wen2024autodroid,king2024sasha,shen2025autoiot,zhang2023appagent}, an in-the-wild proactive agent must infer user needs from massive sensor streams in advance without direct instructions.
Although recent work has explored LLMs and visual LLMs (VLMs)~\cite{lin2024vila} for general-purpose sensor interpretation~\cite{xu2024penetrative,han2024onellm,yu2025sensorchat,han2024onellm,tian2025dailyllm}, delivering timely and content-appropriate proactive assistance requires extracting intention-related cues from these streams.
Extracting such proactive-oriented cues from multi-dimensional sensory data remains challenging.
\textit{Third}, unlike short task episodes or predefined proactive assistance scenarios~\cite{yang2025socialmind,li2025satori,lee2025sensible}, in-the-wild proactive assistance spans daily life, where frequent or unnecessary assistance can easily become intrusive. The agent must recognize when assistance is aligned with user intent and substantively informative, posing challenges for proactive reasoning in the wild.

In this paper, we present \workname, an end-to-end proactive LLM agent system that harnesses sensory contexts on demand to provide in-the-wild proactive assistance throughout daily life. \workname~first adopts an on-demand tiered perception strategy that coordinates always-on low-cost sensing with on-demand high-cost visual perception, enabling continuous perception during in-the-wild use while reducing system overhead.
It then constructs proactive-oriented hierarchical contexts by integrating multimodal sensory streams with persona cues, enabling the agent to reason about the user’s current situation while incorporating longer-term preferences.
Based on these contexts, \workname~employs a VLM-based proactive reasoner to infer user needs and generate proactive assistance decisions, including assistance timing, response content, and tool calling. Finally, \workname~incorporates proactive delivery control and user feedback to reduce unnecessary interruptions and adapt assistance to user preferences over time.
We summarize the contributions of this work as follows.

\begin{itemize}[leftmargin=*]
\item 
We present \workname, an end-to-end proactive agent system that supports in-the-wild assistance in everyday life by continuously perceiving sensory contexts from mobile devices and integrating them with LLM reasoning.

\item
We develop an on-demand tiered perception mechanism to continuously capture multimodal sensory data in the wild with low overhead on mobile devices, and a proactive-oriented context extraction approach that derives hierarchical contexts from massive sensory streams to support proactive reasoning.

\item
We develop a context-aware proactive reasoner that reasons over hierarchical contexts to infer user intent, predict proactive needs, and determine appropriate tool calls, enabling timely and unobtrusive assistance.

\item
We design and implement \workname~on AR glasses and smartphones with an edge server and validate its effectiveness on both a public dataset and a real-world dataset.
\workname~can achieve up to 27.7\% higher accuracy in proactive prediction, 20.5\% lower false detection over state-of-the-art baselines.
A user study with 20 participants further shows that 85\% of participants were willing to use \workname~in their daily lives.

\end{itemize}

%% file: secs/2_related_works.tex
\section{Related Works}

\begin{table}[t]
\footnotesize
\centering
\setlength{\tabcolsep}{3.2pt}
\caption{Comparison of \workname~with existing studies on proactive assistants and LLM agent systems. \ding{182}, \ding{183}, \ding{184}, \ding{185}, and \ding{186} denote text, vision, audio, IMU, and GPS, respectively. N.A. indicates the corresponding item is not applicable or not reported.} 
\vspace{-0.6em}
\label{tab:compare}
\resizebox{\columnwidth}{!}{
\begin{tabular}{lcccccccc}
\toprule
Methods &
\begin{tabular}[c]{@{}c@{}}Agent\\Paradigm\end{tabular} &
\begin{tabular}[c]{@{}c@{}}Temporal\\Span\end{tabular} &
\begin{tabular}[c]{@{}c@{}}LLM\\Reasoning\end{tabular} &
\begin{tabular}[c]{@{}c@{}}Tool\\Use\end{tabular} &
\begin{tabular}[c]{@{}c@{}}Persona-\\lization\end{tabular} &
\begin{tabular}[c]{@{}c@{}}Context\\Modalities\end{tabular} &
\begin{tabular}[c]{@{}c@{}}Application\\Type\end{tabular} &
\begin{tabular}[c]{@{}c@{}}System\\Deployment\end{tabular} \\
\midrule

CybreMinder~\cite{dey2000cybreminder}
& N.A. & Daily-life & \xmark & \xmark & \xmark & \ding{182} & Reminders & Desktop \\

AutoDroid~\cite{wen2024autodroid}
& Reactive & Episodic & \cmark & \cmark & \xmark & \ding{182} & Smartphone automation & Phone \\

Proactive Agent~\cite{lu2024proactive}
& Proactive & Daily-life & \cmark & \xmark & \cmark & \ding{182} & Computer use & Desktop \\

SocialMind~\cite{yang2025socialmind}
& Proactive & Episodic & \cmark & \xmark & \cmark & \ding{183}\ding{184} & Social interactions & Glasses \\

ContextAgent~\cite{yang2025contextagent}
& Proactive & Episodic & \cmark & \cmark & \cmark & \ding{183}\ding{184} & Open-ended scenarios & N.A. \\

ProMemAssist~\cite{pu2025promemassist}
& Proactive & Episodic & \cmark & \xmark & \xmark & \ding{183}\ding{184} & Predefined scenarios & Glasses \\

Sensible Agent~\cite{lee2025sensible}
& Proactive & Episodic & \cmark & \xmark & \xmark & \ding{183}\ding{184} & Predefined scenarios & XR headset \\

\rowcolor[gray]{0.9}
\textbf{\workname}
& Proactive & Daily-life & \cmark & \cmark & \cmark & \ding{182}\ding{183}\ding{184}\ding{185}\ding{186} & Open-ended scenarios & Glasses + Phone \\

\bottomrule
\end{tabular}
}
\end{table}

\subsection{LLM Agents for Mobile Systems}
Recent studies extend LLM agents to support task automation in mobile and embedded systems, including mobile UI automation~\cite{wen2024autodroid, lee2024mobilegpt, zhang2023appagent}, embedded code generation~\cite{shen2025autoiot, yang2024embedgenius, englhardt2024exploring}, and sensor system coordination~\cite{gao2024chatiot, liu2024tasking, cheng2024autoiot}.
Several studies leverage screenshots and multimodal LLMs (MLLMs) for UI understanding~\cite{zhang2023appagent,wang2024mobile}, enabling autonomous app navigation and interaction based on explicit user instructions.
Some studies also integrate wearable sensor data with LLM reasoning for personal healthcare assistants~\cite{kim2024health}.
Others explore LLM-driven program synthesis for IoT and embedded systems, where natural-language task requirements are translated into executable programs~\cite{shen2025autoiot,yang2024embedgenius,gao2024chatiot}.
Results work also~\cite{king2024sasha,liu2024tasking} employ LLM agents to interpret user queries and coordinate smart home devices and heterogeneous sensor systems.
Recent multimodal agents such as M3-Agent~\cite{long2025seeing} leverage vision and audio to build long-term memory for question answering tasks.
However, these systems remain reactive LLM agents, relying on explicit user instructions to initiate agent services and lacking the ability to harness rich surrounding sensory context to anticipate user intentions and proactively offer assistance.

\subsection{Proactive Assistants and Agent Systems}

Early proactive assistants were primarily built on rule-based triggering systems~\cite{rhodes1997wearable,dey1999conference,dey2000cybreminder,fall_detection_apple_watch}, user intent modeling~\cite{horvitz1998lumiere}, and task-oriented planning agents~\cite{yorke2009like,myers2007proactive}. 
Their proactivity was typically event-driven, with assistance triggered automatically once predefined conditions were satisfied.
Recent studies have begun to leverage LLMs for more autonomous and open-ended proactive assistance.
For example, Proactive Agent~\cite{lu2024proactive} and FingerTip 20K~\cite{yang2025fingertip} rely on contexts derived from desktop or smartphone interfaces to support proactive behaviors such as detecting prolonged inactivity or offering code completion.
However, their input mainly comes from desktop or smartphone screens, rather than rich multimodal sensory signals from the physical world.
Other studies explore predicting teammates’ actions in multi-agent systems~\cite{zhang2024proagent} and designing re-asking mechanisms to reduce ambiguity~\cite{zhang2024ask}.
Some recent work, such as SocialMind~\cite{yang2025socialmind} and OS-1~\cite{xu2024can}, integrates multimodal sensory context with LLM reasoning to provide proactive, social assistance and companionship.
Recent studies have explored proactive assistants that leverage egocentric video, audio, or other contextual cues to infer user needs and deliver unobtrusive assistance~\cite{pu2025promemassist,lee2025sensible,li2025satori}. However, they primarily focus on short-term episodes or predefined scenarios, rather than continuous, open-ended assistance in everyday life. ContextAgent~\cite{yang2025contextagent} introduces a framework for proactive LLM agents with multimodal sensing and tool use, but primarily focuses on benchmark construction rather than continuous in-the-wild sensing and reasoning on mobile systems.

\subsection{Sensor Understanding and Adaptive Perception}
\myparagraph{Understanding Sensor Contexts via LLMs}.
Recent studies have explored diverse approaches to leveraging LLMs for understanding sensor data.
Penetrative AI~\cite{xu2024penetrative} and HARGPT~\cite{ji2024hargpt} prompt LLMs with domain expertise and raw sensor recordings as demonstrations to perform sensing tasks, such as activity recognition and heart rate detection.
Recent studies, such as
ContextLLM~\cite{post2025contextllm}, AutoLife~\cite{xu2024autolife}, and LLMSense~\cite{ouyang2024llmsense} employ LLMs to perform reasoning on sensor perceptions from specialized sensing models, enabling a comprehensive understanding of sensor data.
Studies such as SensorChat~\cite{yu2025sensorchat}, OneLLM~\cite{han2024onellm}, and OPERA~\cite{zhang2024towards} align large-scale text with sensor data in a unified embedding space,
enabling captioning and reasoning over sensor data.
However, existing studies primarily focus on interpreting sensor data and providing general-purpose descriptions. 
In contrast, \workname~not only understands current multimodal sensor streams, but also performs on-demand sensing, anticipates users’ potential needs, and provides proactive assistance through an LLM agent with tool use.

\myparagraph{Adaptive Perception for Sensor Data.}
Prior studies have explored adaptive sensing and sampling to reduce perception and inference overhead on edge devices~\cite{neseem2020adasense,lu2011speakersense,yan2012energy,li2020reducto}.
AdaSense~\cite{neseem2020adasense} employs a set of predefined accelerometer sampling configurations and dynamically adjusts sampling frequency based on activity stability for low-power activity recognition on wearable devices.
SpeakerSense~\cite{lu2011speakersense} dynamically controls audio sensing based on speech activity, activating speaker identification only when needed to reduce energy consumption.
A3R~\cite{yan2012energy} dynamically switches the accelerometer sampling rate for continuous activity recognition.
EgoTrigger~\cite{paruchuri2025egotrigger} uses audio-based event detection to trigger smart-glasses image capture for energy-efficient egocentric sensing.
Reducto~\cite{li2020reducto} and FilterForward~\cite{canel2019scaling} utilize frame differences and low-level features to enable efficient vision data processing.
However, existing approaches are typically tailored to predefined tasks and rely on fixed cues or heuristic rules, limiting their ability to perform proactive reasoning over open-world sensory contexts.

%% file: secs/3_motivation.tex
\section{Background and Motivation}
\subsection{Application Scenarios}
\workname~is designed for continuous, in-the-wild proactive assistance in everyday life through mobile and wearable devices. Rather than serving only short-term episodes or task-specific situations, it supports open-ended daily scenarios such as commuting, shopping, and dining. 
\workname~continuously captures surrounding multimodal signals, including egocentric video, audio, motion, location, and reasons over the sensory contexts to anticipate user needs, determine appropriate timing, and deliver timely and unobtrusive assistance through smart glasses or audio cues. This allows \workname~to reduce both the physical effort of manually invoking an agent and the cognitive burden of tracking useful information during ongoing activities.
For example, \workname~can recommend suitable dishes or provide dietary reminders while the user is dining out, compare online prices and summarize product ratings while shopping, and present relevant transit information when the user approaches a bus stop.
\workname~can further support broader applications such as assistive systems for people with visual impairments~\cite{yang2024viassist} and healthcare support~\cite{kim2024health,yang2024drhouse}.

\subsection{Motivation and Challenges}
\subsubsection{Limitations of Reactive Agents.}

Existing studies primarily focus on reactive LLM agents~\cite{wen2024autodroid,shen2025autoiot,lee2024mobilegpt,long2025seeing,king2024sasha}, which require explicit user instructions to initiate tasks. Although these agents can automate user requests, their reactive nature still imposes substantial physical and cognitive burden. First, users must manually invoke the agent whenever assistance is needed, such as by unlocking a phone, opening an interface, and issuing a request. Such interactions remain cumbersome during ongoing activities such as walking, meetings, or social interactions, where attention is already occupied. Second, assistance is available only when users explicitly recognize a need and decide to ask for help. In many daily situations, users may overlook opportunities for agent or tool support when their attention is divided. These limitations motivate a shift from reactive agents to proactive agents that can anticipate user needs and provide timely, unobtrusive support.

\begin{figure}[t!]
    \centering
    \begin{subfigure}[t]{0.48\columnwidth}  
        \centering
        \includegraphics[width=1\linewidth]{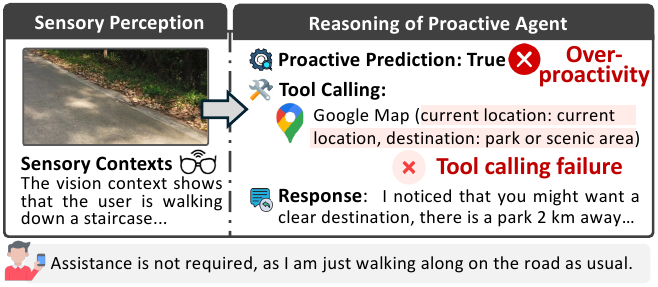}
        \vspace{-1.5em}
        \caption{Over-proactivity and tool errors.}
        \label{fig:motivation_overproactive}
    \end{subfigure}
    \hfill
    \begin{subfigure}[t]{0.51\columnwidth}  
        \centering
        \includegraphics[width=1\linewidth]{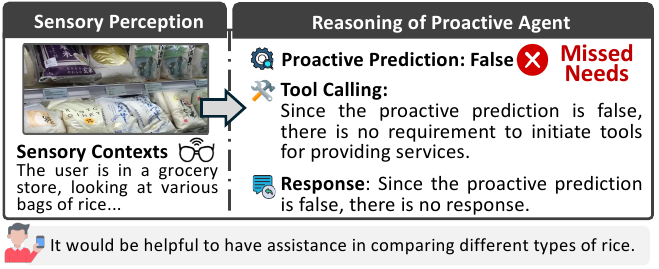}
        \vspace{-1.5em}
        \caption{Missed proactive needs.}
        \label{fig:motivation_missed}
    \end{subfigure}
    \vspace{-1.2em}
    \caption{Motivating examples of adapting existing LLM and VLM agents for in-the-wild proactive reasoning. Existing approaches may generate unnecessary assistance, call inappropriate tools, or miss moments where proactive assistance is needed.}
    \label{fig:motivation_examples}
    \vspace{-.5em}
\end{figure}

\begin{figure}[t!]
\centering

\begin{minipage}[t]{0.49\columnwidth}
    \vspace{0pt}
    \centering

    \begin{minipage}[t]{0.49\linewidth}
        \vspace{0pt}
        \centering
        \includegraphics[width=\linewidth]{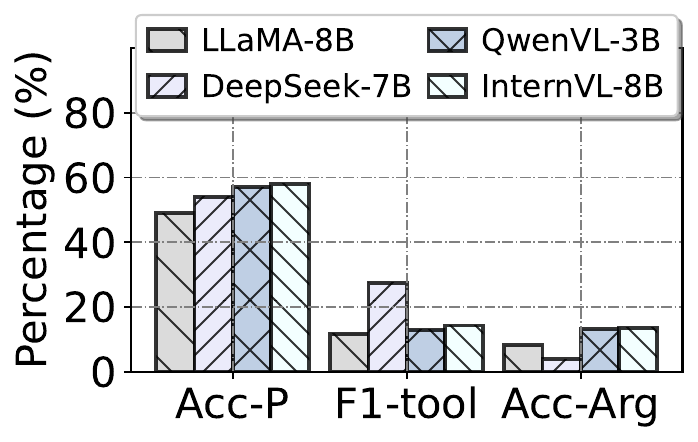}
        \vspace{-2em}
        \captionof{figure}{Adapting existing LLMs and VLMs to the proactive agents.}
        \label{fig:motivation_contextllm}
    \end{minipage}\hfill
    \begin{minipage}[t]{0.49\linewidth}
        \vspace{0pt}
        \centering
        \includegraphics[width=\linewidth]{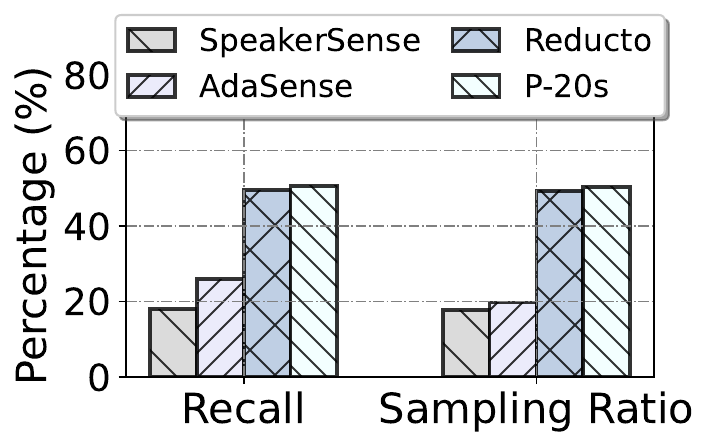}
        \vspace{-2em}
        \captionof{figure}{Existing adaptive perception methods in proactive agent systems.}
        \label{fig:motivation_rule}
    \end{minipage}
\end{minipage}\hfill
\begin{minipage}[t]{0.49\columnwidth}
    \vspace{0pt}
    \centering

    \refstepcounter{figure}
    \label{fig:motivation_vision}
    \setcounter{subfigure}{0}

    \begin{subfigure}[t]{0.48\linewidth}
        \vspace{0pt}
        \centering
        \includegraphics[width=\linewidth]{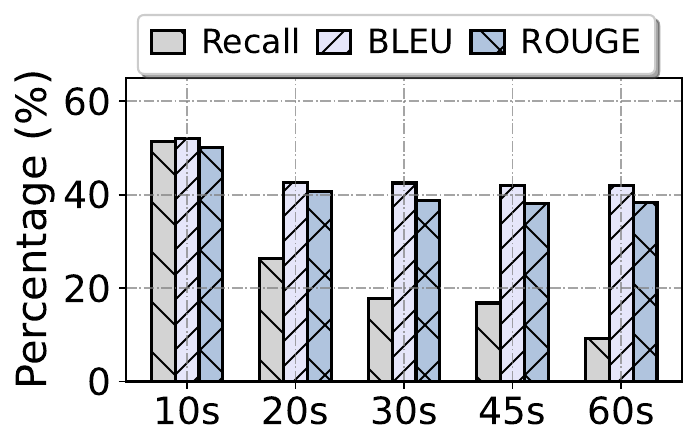}
        \vspace{-1.5em}
        \caption{Recall and caption quality.}
        \label{fig:motivation_vision_sample}
    \end{subfigure}\hfill
    \begin{subfigure}[t]{0.48\linewidth}
        \vspace{0pt}
        \centering
        \includegraphics[width=\linewidth]{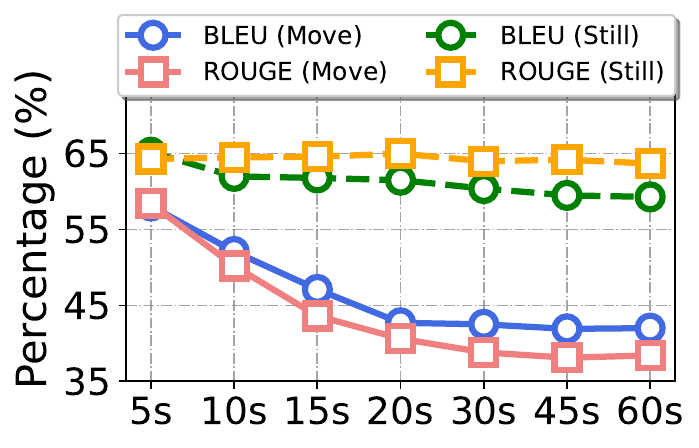}
        \vspace{-1.5em}
        \caption{Impact of scene variation.}
        \label{fig:motivation_caption_still_move}
    \end{subfigure}

    \vspace{-1em}
    \captionof*{figure}{Figure~\thefigure: Impact of egocentric video on proactive agent reasoning. The X-axis denotes the periodic sampling interval.}
\end{minipage}

\vspace{-.5em}
\end{figure}

\subsubsection{Proactive Reasoning in the Wild}
The limitations of existing reactive agents motivate a shift toward proactive agents that can infer user needs and provide timely assistance without explicit instructions. However, existing studies mainly focus on desktop~\cite{lu2024proactive} and smartphone~\cite{yang2025fingertip}, where perception is confined to on-screen content, or on short-term episodes in predefined tasks~\cite{pu2025promemassist,lee2025sensible}, rather than continuous assistance throughout daily life.

To examine whether existing LLM and VLM agents can support such assistance, we conduct experiments on the public dataset CAB-Lite~\cite{yang2025contextagent}, which contains multimodal sensory data with proactive assistance annotations.
We adapt reactive agents to infer user needs from sensory contexts rather than from explicit user instructions.
Agents are required to determine whether proactive assistance is needed and call appropriate tools to assist. For LLM agents, we additionally employ a separate VLM for visual captioning, and use in-context learning (ICL)~\cite{dong2022survey} with ten examples to adapt them for proactive reasoning.
We evaluate proactive assistance triggering using \textit{Acc-P}, and assess tool-calling performance using \textit{F1-tool} and \textit{Acc-Args}~\cite{yang2025contextagent}.
Figure~\ref{fig:motivation_examples} shows examples of adapting existing LLM and VLM agents for in-the-wild proactive reasoning in everyday scenarios, such as walking outdoors and shopping.
Figures~\ref{fig:motivation_examples} and~\ref{fig:motivation_contextllm} show that they often fail to provide assistance at appropriate moments, resulting in both excessive proactivity and missed needs, while inaccurate tool calls further reduce the usefulness of the assistance.
Results highlight the challenges of in-the-wild proactive assistance, including extracting proactive-relevant cues from sensory streams and mapping them to assistance that is unobtrusive and substantively informative.

\subsubsection{Overhead of Continuous In-the-Wild Assistance.}
Proactive assistance in the wild requires continuous perception to provide timely support. In particular, egocentric vision is critical for recognizing cues related to user intent.
However, always-on high-rate visual perception is costly for resource-constrained wearable devices such as smart glasses.
We first examine the role of egocentric video in proactive reasoning using real-world data annotated with proactive service needs. Specifically, we evaluate proactive need detection and video caption quality under different video sampling rates. We use \textit{recall} to measure whether moments requiring proactive service are captured, and BLEU~\cite{papineni2002bleu} and ROUGE~\cite{lin2004rouge} to compare caption similarity between 1-second sampling and lower-rate sampling. Figure~\ref{fig:motivation_vision_sample} illustrates that reducing the sampling rate significantly lowers both \textit{recall} and caption quality, highlighting the importance of egocentric video for recognizing intent-related cues in proactive assistance.
Next, we analyze the impact of vision sampling rates when users either remain still or move through changing environments.
Figure~\ref{fig:motivation_caption_still_move} demonstrates that low sampling rates have minimal impact when users are stationary but can miss critical events in dynamic environments.
This highlights the potential of incorporating diverse low-cost sensor data, such as location and motion, to guide high-cost vision sampling, reducing system overhead while maintaining proactive service quality.
We further evaluate strategies for adaptive sampling of vision data.
Specifically, we evaluate AdaSense~\cite{neseem2020adasense} and SpeakerSense~\cite{lu2011speakersense}, which use always-on IMU and audio to trigger vision sampling when key events such as movement and conversation are detected.
In addition to \textit{recall}, we measure the sampling ratio relative to 10s periodic sampling to evaluate efficiency.
We also present the performance of 20s periodic sampling (P-20s) and examine the vision-only input filtering strategy Reducto~\cite{li2020reducto}.
Figure~\ref{fig:motivation_rule} demonstrates that, although they can filter out unnecessary video data, they cannot capture the rich information in multi-modal sensor data to provide proactive assistance correctly, highlighting the challenge of triggering high-cost visual sampling at the right moments for proactive assistance in the wild.

%% file: secs/4_system.tex
\section{System Design}
\subsection{System Overview}
\begin{figure*}
  \centering
\includegraphics[width=1\linewidth]
{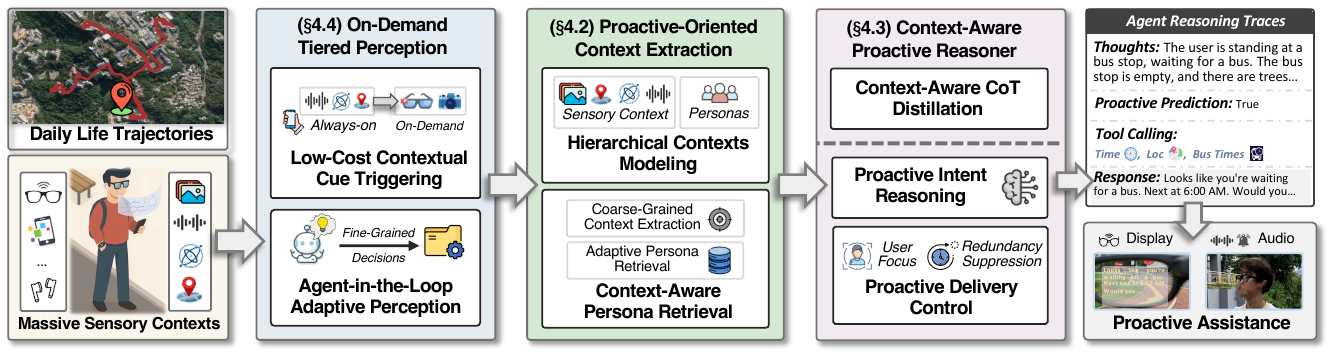}
\vspace{-2.em}
  \caption{System overview of \workname. \workname~performs on-demand tiered perception to continuously capture sensory data in open-world daily life with minimal overhead, extracts proactive-oriented contexts that include short-term sensory contexts and long-term user preferences, and feeds them into a proactive reasoner to deliver proactive assistance.}
  \vspace{-1.em}
\label{fig:system_overview}
\end{figure*}

Figure~\ref{fig:system_overview} illustrates the system overview of \workname.
\workname~first employs an on-demand tiered perception mechanism to continuously capture the user’s surrounding context via multimodal sensors (\S~\ref{sec:Active On-Demand Tiered Perception}).
It coordinates always-on low-cost sensors with on-demand high-cost sensors, adaptively adjusting their sampling rates to ensure efficient perception while capturing proactive-related cues.
Next, \workname~employs a proactive-oriented context extraction approach that derives hierarchical contexts from the massive sensor data (\S~\ref{sec:context_extraction}), spanning short-term sensory contexts and long-term user preferences.
Finally, \workname~employs a context-aware proactive reasoner based on VLMs to reason over these contexts, determine whether proactive assistance is needed, and deliver unobtrusive assistance through proactive delivery control (\S~\ref{sec:Context-aware Proactive Reasoning}).

\subsection{Proactive-Oriented Context Extraction}
\label{sec:context_extraction}
Although existing studies have explored using LLMs or VLMs to interpret sensor data~\cite{han2024onellm,post2025contextllm,xu2024autolife,yu2025sensorchat}, they primarily focus on general sensor signal understanding rather than inferring user intents and proactive needs, which limits their ability to support proactive agent systems.
Moreover, directly reasoning over massive sensory contexts with substantial irrelevant information may hinder proactive prediction and increase system overhead~\cite{shi2023large,liu2024lost}.
To address this challenge, we develop a proactive-oriented context extraction approach that derives short-term sensory contexts from massive sensor data while incorporating long-term user preferences for proactive reasoning.

\subsubsection{Hierarchical Context Modeling} 
\label{sec:Hierarchical Context Structuring}
Recent proactive agent systems primarily focus on users’ current multimodal context or short-term cognitive state~\cite{lee2025sensible,pu2025promemassist}. In contrast, \workname~derives hierarchical contexts capturing both the current sensory situation and longer-term user preferences for proactive reasoning in daily life.

\noindent\textbf{Sensory Contexts.}
\workname~continuously perceives multi-modal sensor streams, including egocentric video, audio, motion, and location, to derive sensory contexts that capture both the surrounding environment and the user’s current state.
Specifically, \workname~utilizes GPS coordinates together with reverse geocoding to identify nearby points of interest, such as bus stops and supermarkets, as well as their proximity, which form the location context. It further uses IMU signals to infer the user’s motion state, such as static or moving, as the motion context. In addition, \workname~utilizes audio to detect whether conversations occur~\cite{SileroVAD}, as such interactions often call for proactive services, and transcribes speech into dialogue as the audio context.
\workname~also maintains a short-term history for temporal reasoning through a context window of length $L$, which stores recent records of sensory contexts and proactive responses with their timestamps (see \S~\ref{sec:Context-aware Proactive Reasoning}).

\noindent\textbf{Persona Contexts}.
Since users may differ in their preferences and needs for proactive assistance, even under the same sensory contexts, the appropriate timing and content of assistance can vary substantially.
Although prior work~\cite{li2025satori} personalizes proactive assistance by modeling users’ beliefs, desires, and intentions, it mainly focuses on immediate states in predefined scenarios.
In contrast, \workname~incorporates persona contexts to capture longer-term, context-aware preferences in open-world daily life. In particular, \workname~allows users to provide personas in natural language, denoted as $\mathbf{P}_{\text{init}}$, such as ``\textit{I am price-sensitive},'' ``\textit{I prefer taking the metro or campus shuttle},'' and ``\textit{I care about diet and healthy eating}.''
These persona contexts are represented as textual descriptions of user preferences and traits. In addition to user-provided personas, \workname~also supports context-aware persona extraction from subsequent user feedback, as described in \S~\ref{sec:Context-Aware Persona Updating}.

\subsubsection{Context-Aware Persona Updating}
\label{sec:Context-Aware Persona Updating}
In addition to initially provided personas, \workname~also infers context-aware personas from user feedback and the associated sensory contexts.
Although recent studies have explored persona extraction~\cite{gao2024aligning,ramos2024transparent,zhong2024memorybank}, they primarily infer personas from textual interactions, such as chatbot conversations.
In contrast, \workname~derives context-aware personas by jointly modeling users’ verbal feedback and surrounding sensory context, capturing situation-specific preferences across daily scenarios.

Specifically, \workname~records the current sensory context (e.g., timestamps and location), the delivered proactive assistance, and the user’s verbal feedback as an interaction record \(r_i=(c_i, a_i, f_i)\), where \(c_i\), \(a_i\), and \(f_i\) denote the sensory context, proactive assistance, and verbal feedback, respectively.
Over time, these records accumulate into an interaction history \(\mathcal{H}\).
\workname~uses an LLM to infer personas from \(\mathcal{H}\), denoted as \(\mathbf{P}_{\text{inf}}=\texttt{LLM}(p_{\text{persona}}, \mathcal{H})\), where \(p_{\text{persona}}\) denotes the persona extraction prompt.
The inferred personas capture users’ context-aware preferences associated with specific situations, locations, and activities.
For example, \textit{``the user does not want schedule reminders at the campus bus stop near GPS coordinates [Lat A, Lon A]''}.
Prompt details are provided in the Appendix.
The prompt also instructs the LLM to distinguish personas across scenarios.
For example, \textit{``during commuting, the user accepts time-critical transit reminders'' and `` during work, the user prefers concise notifications focused on the ongoing task and tends to ignore distracting suggestions.''}
All personas, including \(\mathbf{P}_{\text{init}}\) and \(\mathbf{P}_{\text{inf}}\), are maintained in a scenario-grouped persona database \(\mathcal{D}_{\text{persona}}\), which stores a persona set for each scenario. 
\workname~maintains this database through semantic matching.
Each newly inferred persona is compared against existing entries for similarity and conflict.
Conflicting personas are replaced, while similar ones are merged.

\begin{figure}
\begin{minipage}[b]{0.53\columnwidth}
     \centering
\includegraphics[width=1\linewidth]
{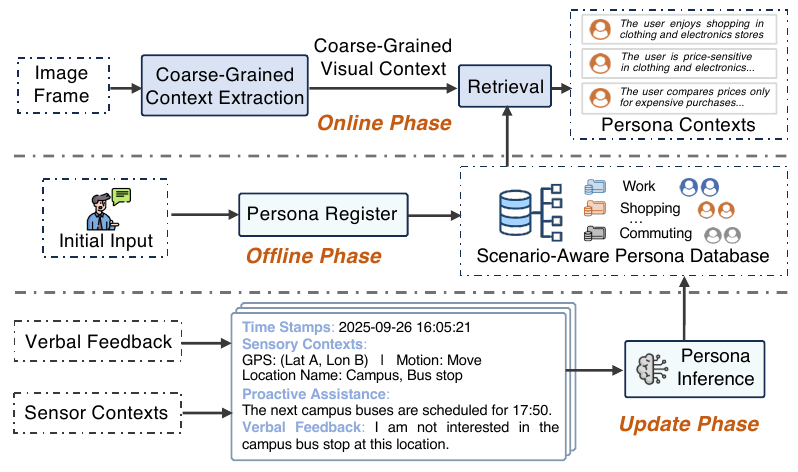}
\vspace{-2.em}
  \caption{Pipeline of context-aware persona retrieval. The upper-right part shows examples of retrieved personas.}
\label{fig:persona_extraction}
\end{minipage}
\hspace{0.01\columnwidth} 
\begin{minipage}[b]{0.43\columnwidth}
     \centering
\includegraphics[width=1.0\columnwidth]{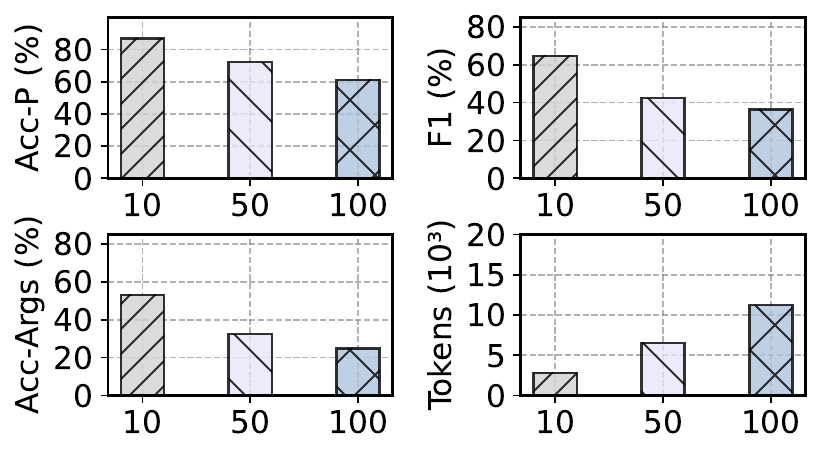}
            \vspace{-2em} 
    \caption{Performance of proactive prediction and system overhead as the number of irrelevant personas increases. \textit{Acc-P} denotes trigger accuracy, while \textit{F1-tool} and \textit{Acc-Args} denote tool-calling performance.} 
    \label{fig:motivation_persona_retrieval}
\end{minipage}
\vspace{-1.3em}
\end{figure}

\subsubsection{Context-Aware Persona Retrieval}
\label{sec:persona retrieval}
An individual may possess multiple personas. 
However, directly incorporating all persona contexts into proactive reasoning may incur substantial system overhead and even reduce the accuracy of user-need prediction~\cite{shi2023large,liu2024lost}.
Figure~\ref{fig:motivation_persona_retrieval} demonstrates that as the number of personas irrelevant to the current scenario increases, relevant personas are overwhelmed in the lengthy prompt, reducing proactive prediction accuracy and leading to higher system overhead.

To address this challenge, we develop a context-aware persona retrieval approach that adaptively integrates personas into proactive reasoning.
Figure~\ref{fig:persona_extraction} shows the workflow.
\workname~first groups user personas by scenario in the offline stage.
At runtime, it extracts coarse-grained visual context and then uses it to retrieve the corresponding scenario-relevant personas.
To obtain visual contexts, a straightforward approach is to employ VLMs to \cmt{generate descriptions and use them to select appropriate personas for proactive reasoning}.
However, this is computationally expensive and would introduce significant latency.
To address this, we develop a coarse-grained context extraction approach to retrieve personas efficiently.
Specifically, \workname~constructs a scenario-object bank $\mathbf{B}$, where each entry $(\mathbf{o}_i, s_i)$ consists of a set of detected objects $\mathbf{o}_i$ paired with a scene category $s_i$.
At runtime, \workname~employs a lightweight visual detection model to extract objects from the current image, which provides a coarse-grained visual context $\mathbf{c}$ to identify the appropriate scenario $s$ based on the scenario-object bank.
To mitigate ambiguity from visually similar scenarios, we retrieve the top-$k$ most similar entries $\mathcal{M}$ from the object-scenario bank $\mathbf{B}$ based on semantic similarity between $\mathbf{c}$ and each entry's object set in the bank as
$
\mathcal{M} = \text{TopK}_{(\mathbf{o}_i, s_i) \in \mathbf{B}} \left( \text{sim}(\phi(\mathbf{c}), \phi(\mathbf{o}_i)) \right),
$
where $\phi(\cdot)$ denotes using a pretrained model to obtain semantic embeddings.
The final scenario $s$ is predicted as the most frequent category among $\mathcal{M}$.
Next, \workname~performs adaptive retrieval by using the predicted scenario \(s\) to retrieve the corresponding persona set, i.e., \(\mathbf{P}_{s}=\mathcal{D}_{\text{persona}}[s]\), where \(\mathcal{D}_{\text{persona}}\) denotes the scenario-grouped persona database.
Only the personas \(\mathbf{P}_{s}\) from the target scenario are incorporated into the reasoning process (\S~\ref{sec:Context-aware Proactive Reasoning}).
\cmt{The scenario categories follow those defined in the CAB-Lite dataset~\cite{yang2025contextagent}.}

\subsection{Context-Aware Proactive Reasoner}
\label{sec:Context-aware Proactive Reasoning}
Proactive agents differ from reactive LLM agents in that they must infer users’ latent needs from multimodal contexts rather than from explicit instructions.
\workname~employs a context-aware proactive reasoner that jointly reasons over sensory and persona contexts to infer user intent and deliver tool-augmented proactive assistance.

\begin{figure}
  \centering
\includegraphics[width=0.9\linewidth]{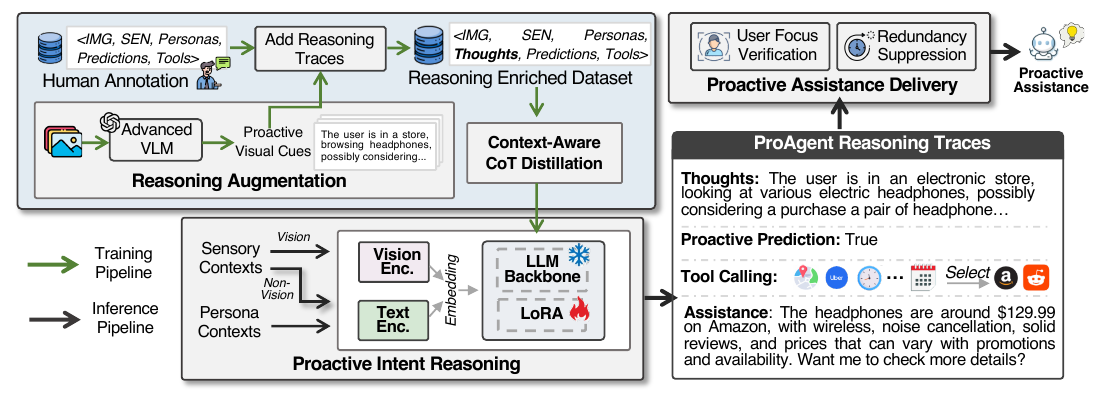}
\vspace{-1em}
  \caption{Illustration of the context-aware proactive reasoner in \workname. `Enc.' represents `Encoder'.}
  \vspace{-1.em}
\label{fig:VLM_reasoning}
\end{figure}

\subsubsection{Proactive Intent Reasoning}
\label{sec:Proactive Intent Reasoning}
This section introduces the inference pipeline of the reasoner.
Unlike prior proactive assistants that first convert vision data into textual descriptions for LLM reasoning~\cite{yang2025contextagent,pu2025promemassist}, \workname~employs a VLM-based reasoner that directly reasons over raw visual inputs and hierarchical contexts described in \S~\ref{sec:context_extraction}.
As shown in Figure~\ref{fig:VLM_reasoning}, the reasoner generates structured outputs, including thought traces, proactive predictions, tool calls, and the final assistance delivered to the user.

\noindent\textbf{Reasoner Inputs.}
\workname~leverages a unified VLM-based reasoner that jointly processes raw visual data and hierarchical contexts from \S~\ref{sec:context_extraction}. Specifically, raw visual data are fed into the visual encoder, while the remaining contexts, including location, motion, audio, and personas, are sent to the text encoder.

\noindent\textbf{Context-Aware CoT Reasoning.}
To avoid black-box predictions, we equip the reasoner with a think-before-acting mechanism.
The reasoner first generates an explicit description of the current visual inputs using Chain-of-Thought (CoT)~\cite{wei2022chain}, e.g., \textit{``the user is in a store, looking at various headphones, possibly considering\ldots''}.
Our experimental results demonstrate that this step allows the reasoner to better understand the current situation than directly mapping inputs to outputs via supervised learning, and improves proactive prediction performance.

\noindent\textbf{Proactive Predictions.}
Next, the reasoner determines whether the current contexts warrant proactive assistance. If no proactive need is identified, ProAgent keeps silent and leaves the user undisturbed. If proactive assistance is needed, the reasoner further generates a tool-use plan to support the subsequent assistance.

\noindent\textbf{Tool Calling for Proactive Agent.}
Once \workname~determines that the user requires proactive assistance, it calls external tools, such as transit schedules and online price comparison services, to assist the user.
Finally, \workname~integrates sensory and persona contexts, thoughts, and tool results to generate the final assistance.
Details of the proactive reasoner prompt are provided in the Appendix.

\subsubsection{Context-Aware CoT Distillation.}
\label{sec:Context-Aware CoT Distillation}
This section introduces the training pipeline of the reasoner.
Although in-context learning (ICL)~\cite{dong2022survey} can adapt MLLMs to proactive reasoning tasks, integrated examples incur additional inference overhead. 
Additionally, directly applying supervised fine-tuning (SFT)~\cite{achiam2023gpt} on sensor data and proactive labels leads to limited understanding of the current situation and user intent.

To address this, we develop a context-aware CoT distillation approach to fine-tune the VLM reasoner.
Specifically, we first use a human-annotated dataset $\mathcal{D}$, where each sample contains $\texttt{<}$\textit{img, sen, personas, predictions, tools}$\texttt{>}$. Here, \textit{img}, \textit{sen}, and \textit{personas} denote raw images, non-visual sensory contexts (e.g., location, motion, audio, and history), and user personas, respectively. \textit{predictions} and \textit{tools} denote the corresponding proactive predictions and tool calls.
To enable the VLM to think before acting, i.e., first reason about the current situation and user intent before predicting proactive needs and tool calls, we employ an additional advanced VLM with a proactive-oriented prompt to generate descriptions for each image. 
Details of the prompt are provided in the Appendix.
Next, we use the generated descriptions as reasoning traces to augment $\mathcal{D}$, yielding a reasoning-enriched dataset $\widetilde{\mathcal{D}}$ in which each sample is $\texttt{<}$\textit{img, sen, personas, thoughts, predictions, tools}$\texttt{>}$, where \textit{thoughts} denotes the reasoning traces for the current visual context.
This encourages the VLM to first generate interpretable thoughts about the current context rather than directly predicting proactive need and tool calls.
Finally, we fine-tune the VLM on $\widetilde{\mathcal{D}}$ and apply Low-Rank Adaptation (LoRA)~\cite{hu2022lora} to reduce training cost.

\begin{figure}[t]
    \centering
    \begin{subfigure}{0.33\columnwidth}
        \centering
        \includegraphics[width=1\columnwidth]{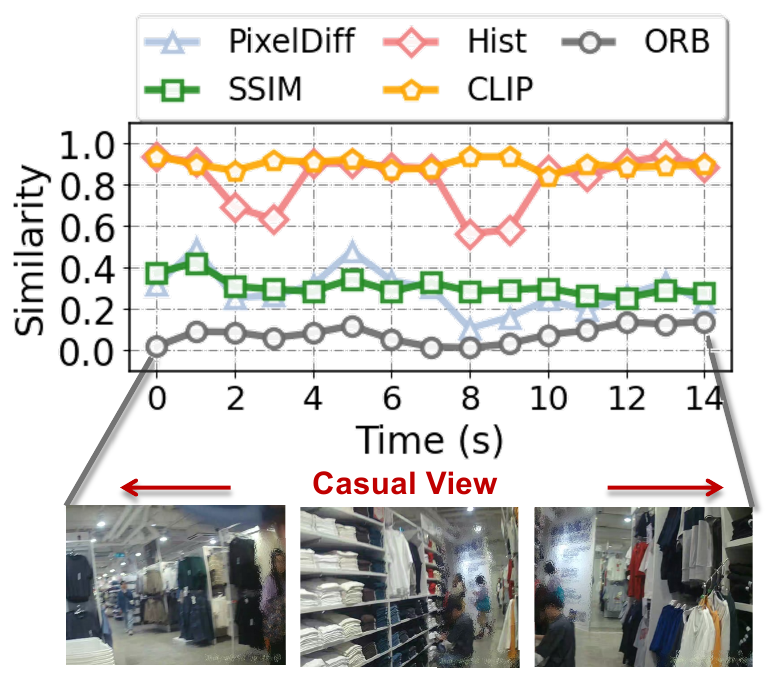}
     \vspace{-1.5em}
        \caption{Casual viewing.} \label{fig:motivation_casual}
    \end{subfigure}
    \begin{subfigure}{0.325\columnwidth}  
        \centering \includegraphics[width=1.0\columnwidth]{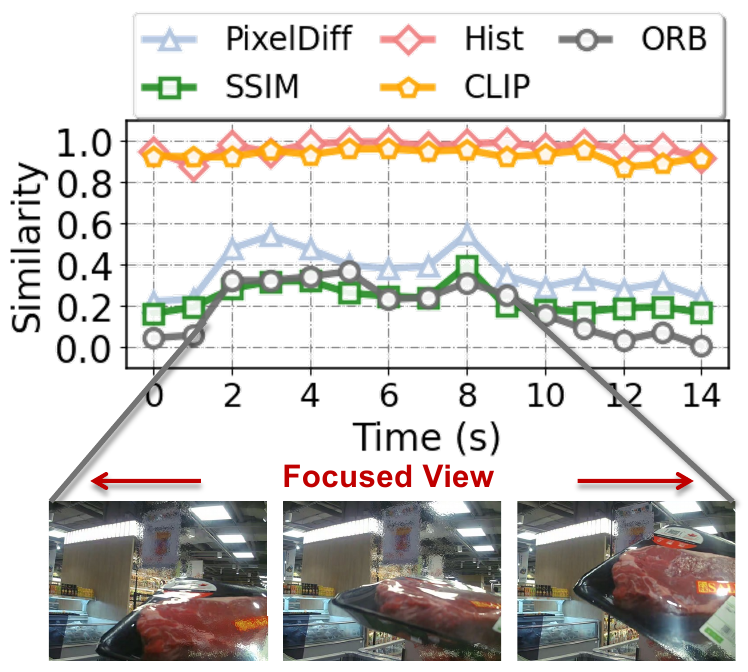}
      \vspace{-1.5em}
        \caption{Focused viewing.} \label{fig:motivation_focus}
    \end{subfigure}
    \begin{subfigure}{0.32\columnwidth}  
        \centering \includegraphics[width=1.0\columnwidth]{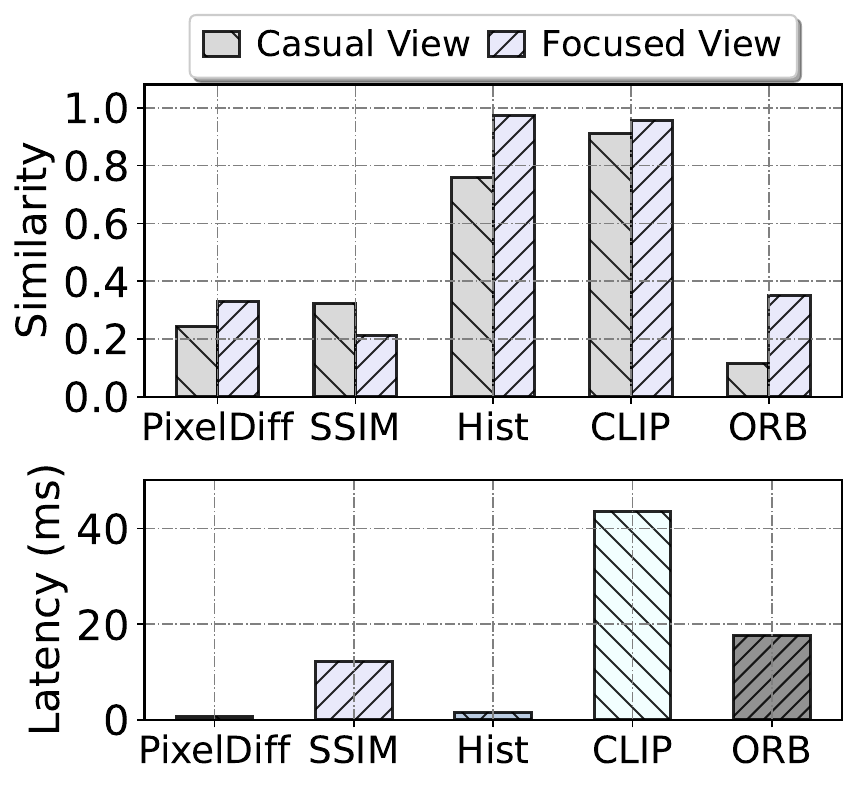}
      \vspace{-1.5em}
        \caption{Similarity and latency.} \label{fig:draw_bar_similarity_latency}
    \end{subfigure}
    \vspace{-1.2em}
    \caption{
The left figure illustrates casual viewing, where continuous motion leads to large temporal variations. The middle figure shows focused viewing, where attention to a specific object results in higher temporal stability. The right figure shows the similarity differences between casual and focused viewing for different metrics, along with their per-frame latency.}
\label{fig:motivation_casual_focus}
      \vspace{-1.em}
\end{figure}

\subsubsection{Proactive Delivery Control}
\label{sec:Proactive Assistance Delivery}
Proactive assistance requires not only accurate intent inference, but also decisions about when inferred assistance should be delivered.  
Even when the inferred assistance is reasonable, directly presenting every candidate response may still cause unnecessary interruptions in daily life. In egocentric streams, brief visual exposure does not always indicate genuine user interest, and prolonged exposure to similar contexts may repeatedly trigger near-duplicate assistance.
To address this challenge, \workname~employs a proactive delivery control mechanism to determine whether candidate assistance should be delivered to the user.

\noindent\textbf{User Focus Verification.}
In egocentric streams, relevant visual targets may appear only briefly as the user walks past objects or turns their head. 
Such brief exposure may be mistaken for genuine user interest, leading to unnecessary assistance.
To reduce unnecessary proactive triggers, \workname~employs a lightweight user focus verification mechanism before delivery.
Figure~\ref{fig:motivation_casual_focus} presents a shopping example under casual viewing and focused viewing. When the user briefly passes by a product, proactive assistance is often unnecessary, whereas sustained viewing more likely indicates interest. This difference is reflected in the visual consistency across consecutive frames.
We compare several representative similarity measures, including pixel difference (PixelDiff), histogram similarity (Hist), SSIM~\cite{wang2004image}, semantic difference (CLIP~\cite{radford2021learning}), and ORB matching~\cite{rublee2011orb}. Figure~\ref{fig:draw_bar_similarity_latency} demonstrates that ORB achieves the largest similarity gap between casual and focused viewing while incurring substantially lower latency than CLIP. Therefore, we adopt ORB match consistency for focus verification.
Specifically, \workname~detects and matches ORB keypoints across consecutive frames and computes an ORB match ratio $m_t$. A candidate response is delivered only when $m_t>\tau_f$, where $\tau_f$ is the focus threshold. High match consistency suggests sustained attention rather than a brief pass-by observation.

\noindent\textbf{Temporal Redundancy Suppression.}
Repeatedly delivering semantically similar assistance in consecutive moments can be unnecessary and intrusive. \workname~therefore measures the semantic similarity between a newly generated response and recent responses using BERT~\cite{devlin2019bert}, and suppresses the response when the similarity exceeds a redundancy threshold $\tau_t$, thereby reducing repeated proactive assistance.

\subsection{On-Demand Tiered Perception}
\label{sec:Active On-Demand Tiered Perception}
Unlike reactive LLM agents that are activated only upon explicit user queries, proactive agent systems must continuously perceive surrounding sensory contexts.
This introduces additional challenges for mobile devices such as smart glasses, especially when capturing high-cost visual data.
However, existing adaptive sampling approaches are often task-specific, relying on predefined cues or heuristic rules~\cite{neseem2020adasense,lu2011speakersense,yan2012energy,paruchuri2025egotrigger}, limiting their ability to support LLM-based proactive assistance in continuous, open-world scenarios, as illustrated in Figure~\ref{fig:motivation_rule}.
To address these challenges, we develop an on-demand tiered perception approach that integrates low-cost sensory cues with agent-in-the-loop adaptive perception for selective visual sampling.

\subsubsection{Low-Cost Contextual Cue Triggering}
\label{sec:Low-Cost Contextual Cue Triggering}
Our preliminary results in Figure~\ref{fig:motivation_vision_sample} show that egocentric vision is crucial for accurately understanding user intent for proactive assistance, yet costly to capture continuously on edge devices.
Therefore, \workname~employs an adaptive visual perception strategy that continuously leverages always-on low-cost sensor streams to track contextual changes, while activating vision sampling only on demand.

Specifically, \workname~first employs always-on low-cost sensors to continuously capture coarse contextual cues with minimal overhead. 
In this work, we consider two types of cues: motion cues and POI cues. Motion cues capture changes in the user’s movement state, which often indicate changes in the surrounding context and a higher need for timely perception. POI cues capture nearby places where proactive assistance is more likely to be useful, such as stores, transit stations, restaurants, and museums.
Based on these cues, \workname~adopts a dual-mode visual sampling strategy that switches between low-rate and high-rate vision sampling. By default, \workname~remains in the low-rate mode to reduce overhead. It switches to the high-rate mode when the user is moving or near relevant POIs, as these moments either involve rapid contextual changes or present stronger opportunities for proactive assistance.
Based on our preliminary results in Figure~\ref{fig:motivation_caption_still_move}, we set the high- and low-rate sampling intervals to 5 s and 60 s, respectively, to capture proactive opportunities during movement while reducing unnecessary perception during stable periods.

\subsubsection{Agent-in-the-Loop Adaptive Perception}
The low-cost contextual cues in \S~\ref{sec:Low-Cost Contextual Cue Triggering} provide only a coarse-grained indication of whether to activate high-rate vision sampling.
However, relying solely on fixed cues or heuristics~\cite{neseem2020adasense} to control visual sampling rates supports only limited scenarios and may misinterpret user intent or miss proactive opportunities.
For example, movement can occur both during casual walking with no assistance needed and during in-store browsing, where timely assistance could support decision-making, making motion cues alone insufficient for deciding the visual sampling rate.
To address this, we develop an agent-in-the-loop adaptive perception approach that enables finer-grained sampling decisions through agent reasoning.

\noindent\textbf{Fine-Grained Decisions with Agent Reasoning}.
Under both low- and high-rate sampling modes, visual frames and sensory contexts are sent to VLM reasoner for proactive reasoning (\S~\ref{sec:Context-aware Proactive Reasoning}). The reasoner first predicts whether proactive assistance is needed at the current moment. \workname~then uses this prediction as a lightweight signal for subsequent visual sampling: moments that require proactive assistance suggest that continued high-rate observation may be useful, while moments without proactive need allow the system to return to low-rate sampling.
For example, when the user is browsing products in a store, the reasoner may indicate that proactive assistance is likely useful, and \workname~maintains high-rate sampling for the following moments. In contrast, when the user is simply walking down the street, the reasoner may indicate no proactive need, allowing \workname~to reduce the sampling rate. 
This enables \workname~to refine coarse cue-based triggers through reasoning outputs, without requiring an additional model for sampling control.

\noindent\textbf{Adaptive Sampling Scheduler}.
\workname~further employs an adaptive sampling scheduler to coordinate low-cost cue triggers with agent-based sampling decisions. Low-cost cues can activate high-rate sampling, while the agent's decision can either maintain high-rate sampling or return the system to the low-rate mode.
However, allowing both mechanisms to directly control mode transitions can create conflicts. 
For example, during continuous walking, motion cues may repeatedly reactivate high-rate sampling even after the agent has determined that continued monitoring is unnecessary. 
To reduce such conflicts, \workname~introduces a suppression interval $\tau$. After a cue activates high-rate sampling, repeated activations from the same cue are temporarily suppressed for $\tau$.

%% file: secs/5_evaluation.tex
\section{Evaluation}

\subsection{System Implementation}
\label{sec:implementation}

\subsubsection{Prototype Testbed}
We implement \workname~on two AR glasses platforms, RayNeo X3 Pro~\cite{rayneo} and Rokid~\cite{rokid}, with a Google Pixel 7 smartphone and a back-end server as the prototype testbed.
The Rokid glasses adopt a Qualcomm Snapdragon AR1 and NXP RT600 dual-chip architecture with a binocular green Micro-LED display, while the RayNeo X3 Pro glasses feature Qualcomm’s Snapdragon AR1 Gen-1 wearable platform.
The Android client, implemented in Kotlin and Java ($\approx$1.2 K LOC), packages sensor data with contextual metadata (e.g., timestamps and request types) and renders returned results in the AR overlay. 
On the server side, we implement the system on heterogeneous platforms for evaluation, including two edge servers (an NVIDIA Jetson Orin and a laptop with RTX 1660 Ti) and two high-performance servers (one with an NVIDIA 4090 GPU and another with 8$\times$~NVIDIA RTX A6000 GPUs). 
LLM/VLM inference runs on the Ollama framework~\cite{ollama} via Python wrappers, and the smart glasses communicate with the server via WiFi and cellular networks over HTTPS.

\subsubsection{Configuration}
\label{sec:Implementation Details}
We fine-tune the VLMs using LoRA with a rank of 8, training for 10 epochs at a learning rate of $5 \times 10^{-4}$.
For coarse-grained context extraction, we employ YOLO-v5~\cite{jocher2022ultralytics}, pretrained on the Objects365 dataset~\cite{shao2019objects365}.
In addition, we set $k=30$, and use \texttt{all-MiniLM-L6-v2}~\cite{reimers2019sentence} for semantic similarity comparison.
We use Google’s Geocoding and Places APIs for reverse geocoding and nearby PoI identification.
The advanced VLM used in context-aware CoT distillation is Qwen2.5-VL-32B~\cite{Qwen2.5-VL}.
We implement the agent using an API-based function calling framework~\cite{li2023api}, with a tool set from the CAB dataset~\cite{yang2025contextagent}.
The tool set contains 20 commonly used tools for real-world assistance, such as product price lookup, public transit schedules, and knowledge search.
The suppression interval $\tau$ is set to 20\,s, the context window length $L$ to 4, and the user focus verification threshold $\tau_f$ to 0.2.
The threshold for the temporal redundancy threshold $\tau_t$ is set to 0.5.
By default, we use the Qwen3-VL-4B as the base model.
We also deploy \workname~on VLMs of different scales, including Qwen3-VL-4B/30B, Gemma3-4B/27B, Qwen2.5-VL-3B/8B/32B~\cite{Qwen2.5-VL}, InternVL2.5-2B/8B~\cite{chen2024internvl}, and LLaVA-1.5-7B~\cite{liu2023llava}.

\subsection{Experimental Setup}

\subsubsection{Dataset} 
\label{sec:dataset}
We comprehensively evaluate \workname~using both a public dataset and a real-world dataset.

\noindent\textbf{CAB-Lite Dataset}.
This is a public dataset~\cite{yang2025contextagent} for context-aware proactive agents.
The dataset comprises 300 samples spanning nine common daily-life scenarios, including shopping, travel, chitchat, work, health, outdoors, cooking, leisure, and others.
Each sample contains multimodal egocentric data, including video and audio, along with user personas and annotations for proactive predictions and tool calls.
The proactive predictions label indicates whether an example should trigger proactive assistance.
Tool calls annotate the expected tool call for LLM agents to provide proactive assistance under the given context.
The toolset in this dataset includes 20 commonly used tools covering real-world assistance scenarios, such as product price lookup, public transit schedules, and knowledge search.
We randomly split the dataset into 60\% for training, using either supervised fine-tuning (SFT) or in-context learning (ICL), and 40\% for testing.

\begin{figure}
  \centering
\includegraphics[width=1\linewidth]
{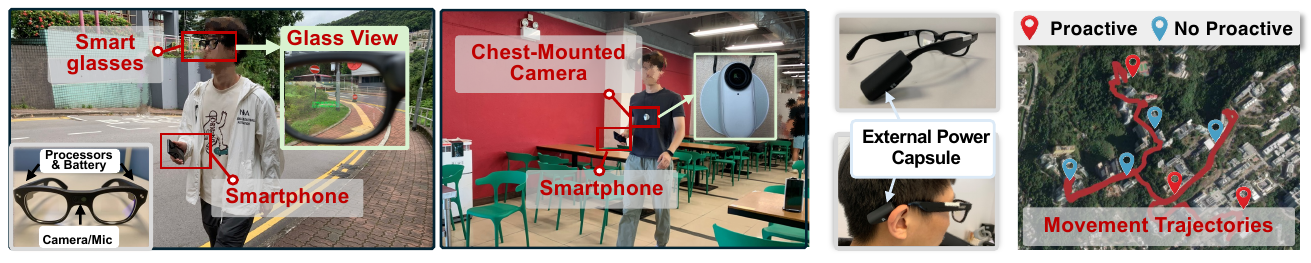}
\vspace{-2em}
  \caption{
Illustration of the real-world testbed and examples of participants’ movement trajectories during data collection. Participants wear egocentric cameras and carry a smartphone to capture multimodal sensor data. }
\label{fig:data_collection}
\vspace{-1em}
\end{figure}

\noindent\textbf{Evaluation on Real-World Testbed}.
While CAB-Lite provides multimodal sensor data paired with annotations for proactive predictions and tool calls, it lacks long-term sensor recordings (e.g., over several hours) and low-cost sensing modalities such as IMU and GPS.
Therefore, we collect an additional in-the-wild dataset for real-world evaluation.
Specifically, we recruited 20 volunteers for real-world data collection, including 12 males and 8 females, with a mean age of 24.3 years.
Participants covered diverse educational backgrounds and daily routines, including undergraduates, postgraduates, and postdoctoral researchers.  
As shown in Figure~\ref{fig:data_collection}, participants recorded egocentric video with either AR glasses or a chest-mounted camera, while carrying a Google Pixel 7 smartphone to collect low-cost sensor data, including audio, IMU, and GPS. 
We also use chest-mounted cameras during data collection, as they can support longer recording sessions.
All videos were recorded at a resolution of 720p.
Participants recorded sensor data across diverse daily-life scenarios, including commuting, work and study, shopping and dining, leisure, travel, and social interactions. Example environments include offices, classrooms, cafeterias, supermarkets, shopping malls, restaurants, libraries, subway systems, bus stops, airports, museums, botanical gardens, hiking trails, coffee shops, and urban outdoor areas. 
Figure~\ref{fig:data_collection} shows an example movement trajectory during data collection.
\textbf{The study was approved by the IRB of the authors’ institution, and informed consent was obtained from all participants.}

\begin{figure}
  \centering
\includegraphics[width=1\linewidth]{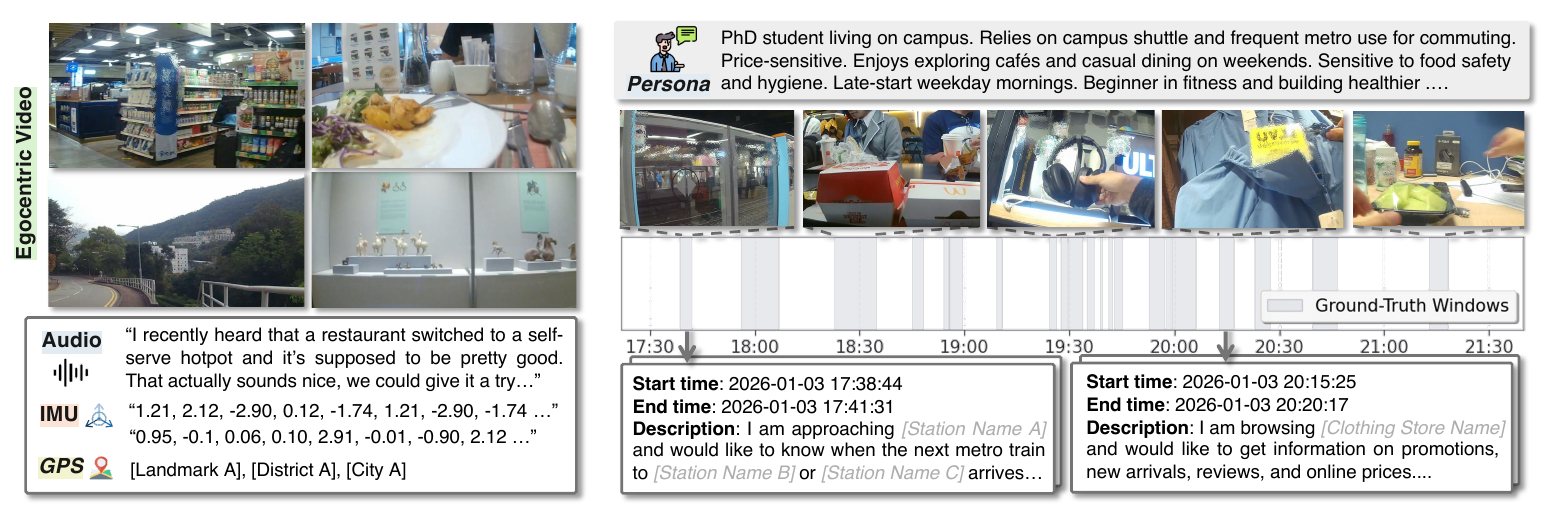}
\vspace{-2em}
  \caption{
Examples of collected multimodal sensor data and human annotations. Participants collect egocentric video, audio, IMU, and GPS data. They then annotate an acceptable time window for each proactive assistance event, specifying its start and end timestamps, expected type, and content, and also provide their personas.}
\label{fig:data_collection_example}
\vspace{-1em}
\end{figure}

After data collection, participants review their recorded videos and annotate proactive assistance, including its timing and service categories.
Specifically, as shown in the right part of Figure~\ref{fig:data_collection_example}, participants annotated an acceptable time window, with start and end timestamps, for each proactive assistance event, together with a free-form description of the expected assistance.
Notably, proactive assistance is considered acceptable within a temporal window rather than at a single precise timestamp. 
To support consistent tool-calling evaluation, the annotations were guided by the service categories in the CAB-Lite annotation schema, which covers 20 common services.
Additionally, each participant provided initial personas. Participants were given example persona statements from prior work~\cite{ge2024scaling} as references, and then wrote a set of short statements describing their own preferences, routines, habits, or assistance needs in daily life, as shown in Figure~\ref{fig:data_collection_example}.
In total, the dataset contains 92 hours of in-the-wild recordings from 20 participants across five cities.

\subsubsection{Evaluation Metrics} 
We extensively evaluate \workname~from multiple perspectives, including the timing of proactive assistance, response quality, tool calling correctness, and system overhead.

\noindent\textbf{Timing Accuracy of Proactive Assistance.}
We first evaluate whether \workname~provides proactive assistance at appropriate moments while avoiding unnecessary interruptions and missed opportunities.
For the real-world dataset, participant-annotated windows serve as ground truth. Each window defines an acceptable interval for delivering proactive assistance, and predictions within these windows are treated as correct.
Following prior studies~\cite{lu2024proactive,yang2025contextagent}, we use \textit{Acc-P} (Proactive Accuracy) to evaluate the overall timing accuracy, \textit{FD} (False Detection) to measure unnecessary proactive triggers, and \textit{MD} (Missed Detection) to measure missed assistance opportunities, respectively.
We also compute \textit{recall}, \textit{precision}, and \textit{F1}-score to evaluate triggering performance.

\noindent\textbf{Response Quality.}
This dimension evaluates the overall usefulness of proactive responses under the current context. 
We adopt an LLM-as-judge approach, which has been validated in prior work~\cite{gu2024survey,chen2024mllm}.
Specifically, we provide the LLM with the participant-annotated assistance need, sensory context, and each method’s prediction, and prompt it to generate a \textit{Usefulness} score based on four aspects: relevance, timeliness, actionability, and non-intrusiveness.
We use GPT-5.2 as the judge model. 
Details of the prompt see Appendix.

\noindent\textbf{Tool-Calling Correctness.}
This dimension evaluates whether \workname~correctly selects and executes the user-intended tools required for proactive assistance.
Following prior works~\cite{basu2024api,yang2025contextagent}, we leverage \textit{F1-tool} to compare tool names between the predicted and ground-truth tool sets, and \textit{Acc-Args} to evaluate whether the agent correctly parses the corresponding tool arguments.
Any errors in the tool selection, API call format, or incorrect parameters are treated as incorrect for \textit{Acc-Args}.

\noindent\textbf{System Overhead.}
This dimension evaluates the system overhead of \workname~in terms of inference latency, communication time, token consumption, and memory usage.
We also evaluate whether the system activates vision sampling and VLM inference when needed.
Specifically, we use \textit{sampling recall} to measure whether high-frequency vision sampling is activated during human-annotated windows, and \textit{sampling ratio} to quantify the frequency of vision sampling and VLM inference relative to dense 1\,second interval sampling.

\subsubsection{Baselines}
To the best of our knowledge, no prior work has developed proactive LLM agent systems that support long-term multimodal sensory perception in the wild.
While some work~\cite{pu2025promemassist} also leverages vision data for proactive assistance, it relies on 1-second interval sampling, which is infeasible for long-term perception over several hours and lacks complementary sensor modalities such as IMU and GPS.
Therefore, we consider two essential components when constructing the baselines: (1) a proactive agent for user-need prediction and (2) an efficient perception strategy for continuous sensing.
The baselines for comparison are as follows.

\noindent\textbf{ContextLLM}.
The original ContextLLM~\cite{post2025contextllm} uses multiple customized models to transform sensor data into descriptive contexts, which are then processed by an LLM for context reasoning. 
We adapt its system prompt to enable it for proactive reasoning, using its intrinsic knowledge to assist users proactively.
Moreover, we equip it with periodic visual sampling strategies set at 10s intervals, and use a VLM to generate visual sensory context.

\noindent\textbf{ContextAgent-ICL}.
This is an ICL-based version of ContextAgent~\cite{yang2025contextagent}, which converts sensor data into context using a VLM and customized models, and employs an LLM-based agent with tool-calling capabilities for proactive prediction.
Moreover, it adopts AdaSense~\cite{neseem2020adasense} to adaptively sample vision data based on IMU-derived motion states, switching to high-frequency sampling when the user is moving and low-frequency sampling otherwise.

\noindent\textbf{ContextAgent-SFT}. 
This baseline approach follows ContextAgent-ICL, but replaces the ICL-based agent with the SFT-based version of ContextAgent~\cite{yang2025contextagent} for proactive prediction. The LLM agent is supervised fine-tuned to map sensor contexts and personas to user-need inference and tool-calling decisions.

The three baselines above follow a two-stage pipeline, where a VLM first generates visual context, followed by an LLM agent for proactive reasoning and tool calling. We further include two baselines that unify both stages within a single VLM agent.

\noindent\textbf{VLM-ICL}.
This approach integrates vision understanding, proactive reasoning, and tool calling into a single VLM with few-shot demonstrations for proactive prediction. In addition, it employs Reducto~\cite{li2020reducto} to eliminate redundant video frames for VLM inference.

\noindent\textbf{VLM-CoT}.
This approach employs Chain-of-Thought~\cite{wei2022chain} prompting to enable the VLM to reason before acting, generating step-by-step reasoning over sensor context and visual data for proactive prediction and tool calling. It also incorporates few-shot demonstrations in the prompt and uses Reducto~\cite{li2020reducto} to filter redundant video frames.

Note that for all baselines employing ICL or CoT strategies, we use five demonstrations from CAB-Lite~\cite{yang2025contextagent} into the system prompts of the VLM/LLM for task adaptation. 
\workname~and all baselines use models of the same scale for both LLM and VLM to ensure a fair comparison.
The high- and low-frequency sampling intervals in AdaSense are set to 5 s and 60 s, respectively.
For Reducto, we use pixel-difference filtering with a threshold of 0.7.

\subsection{Overall Performance}
\label{sec:overall_performance}

\subsubsection{Quantitative Results}
\begin{figure}
  \centering
  \captionsetup{skip=5pt}
\includegraphics[width=0.95\linewidth]{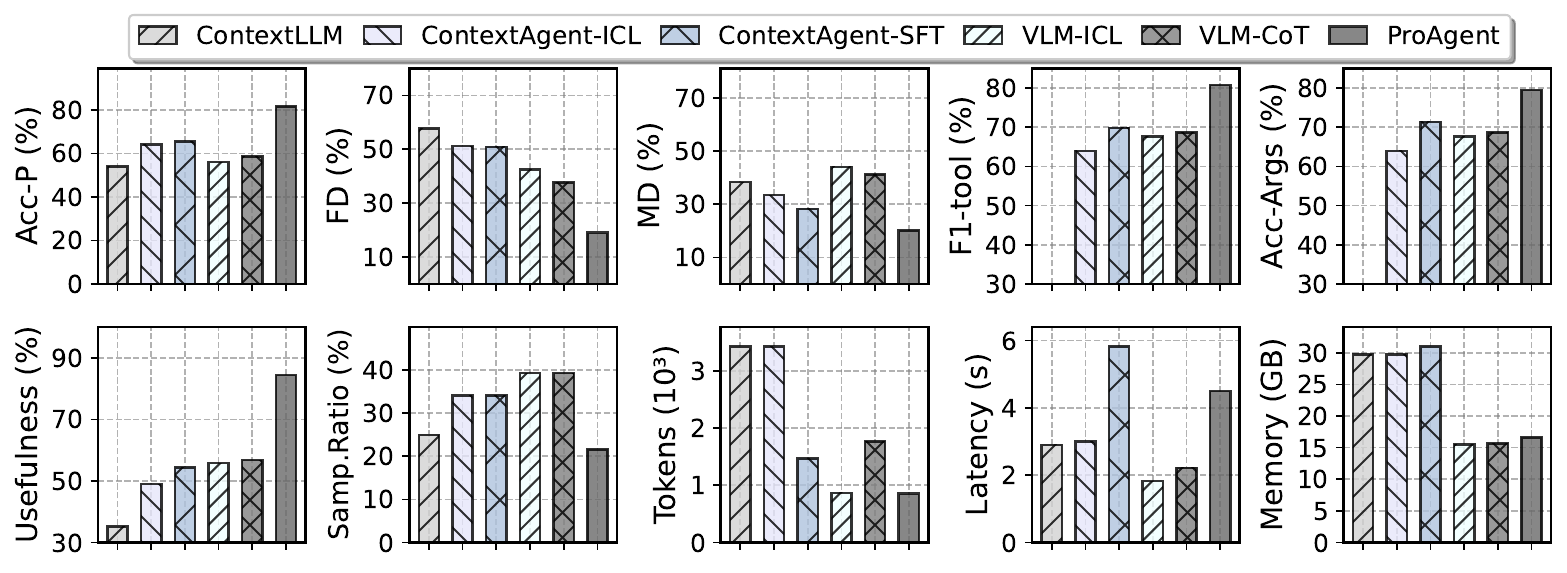}
  \caption{Overall performance of \workname~and baselines on real-world proactive assistance. FD and MD denote false detection and missed detection. Acc-P and Acc-Args denote proactive timing accuracy and argument accuracy in tool calling.}
\label{fig:overall}
\vspace{-1em}
\end{figure}

We first evaluate the overall performance of \workname~on real-world data.
Figure~\ref{fig:overall} illustrates the quantitative results of \workname~and baselines, including the timing accuracy of proactive assistance, response quality, tool-calling accuracy, and system overhead.
Results demonstrate that \workname~consistently outperforms all baselines.
In particular, it achieves up to 27.7\% higher \textit{Acc-P}, 20.5\% lower \textit{FD}, and 9.2\% lower \textit{MD}, showing its effectiveness in triggering proactive assistance at appropriate moments. 
Compared with \workname, the baselines exhibit a tendency toward over-proactive behaviors, leading to higher \textit{FD}.
In addition, \workname~achieves 27.6\% higher \textit{Usefulness} of proactive responses, benefiting from both more accurate proactive triggering and tool use. 
It also achieves 16.8\% higher \textit{F1-tool} and 15.5\% higher \textit{Acc-Args}, indicating more accurate tool calling and argument parsing, which further improves the usefulness of the delivered assistance.
Since ContextLLM relies solely on intrinsic knowledge without tool calling, its \textit{Usefulness}, \textit{F1} and \textit{Acc-Args} remain minimal.

We also assess the system overhead of \workname~during continuous proactive assistance.
Figure~\ref{fig:overall} shows that \workname~requires only 0.86× the sampling ratio of the best baseline. Additionally, compared to two-stage baselines such as ContextAgent-ICL and ContextAgent-SFT, \workname~requires only 0.56× memory and 0.25× tokens, respectively, by unifying vision context extraction and agent reasoning into a single stage. 
\workname~also achieves an average inference time of 4.5 s, validating its ability to provide timely assistance in real-world use.

\begin{figure}
  \centering
\includegraphics[width=1\linewidth]{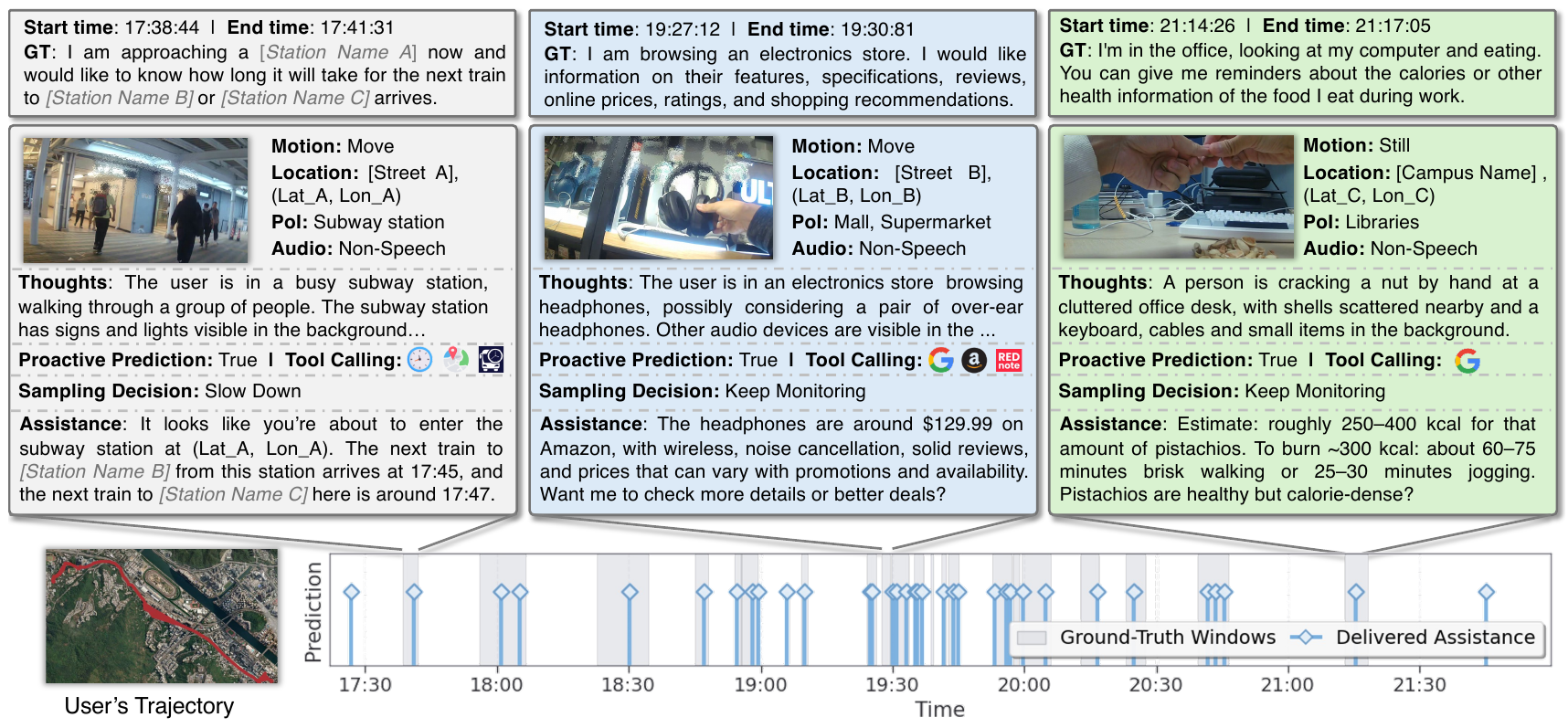}
\vspace{-2em}
  \caption{Examples of \workname’s proactive assistance on real-world data, illustrating sensory contexts, the agent’s thought traces, proactive predictions, tool calling, and final assistance. Ground-Truth windows denote user-annotated acceptable time windows for proactive assistance, and blue markers indicate the moments when \workname~delivers assistance. Specific location names, geographic coordinates, and other identifying information are anonymized for double-blind review.}
  \vspace{-1.5em}
\label{fig:qualitative_results}
\end{figure}

\subsubsection{Qualitative Results}
To better understand \workname's performance, we provide examples of its proactive assistance on real-world data spanning several hours, as illustrated in Figure~\ref{fig:qualitative_results}. We also present the agent’s reasoning process, including perceived sensory context, \workname's thought traces, proactive predictions, tool calling, and final assistance.
Figure~\ref{fig:qualitative_results} demonstrates several cases where \workname~proactively provides assistance, such as approaching a subway station during commuting, browsing products of interest while shopping, and eating snacks during work. 
Results show that most proactive triggers of \workname~fall within user-annotated acceptable time ranges, indicating its ability to accurately anticipate user needs without causing unnecessary interruptions.
By comparison, baselines often exhibit over-proactive behavior, triggering assistance upon detecting relevant objects, such as products, menus, or transit signs, even when these observations do not necessarily indicate user focus or intent.
Baselines also produce missed detections when rule-based or periodic sampling fails to capture important visual cues, preventing assistance from being triggered when needed.
Figure~\ref{fig:qualitative_results} also shows some false detections for \workname, which we discuss in \S~\ref{sec:Proactive Triggering Behavior}.

Additionally, \workname~can call external tools to provide more informative proactive assistance, such as querying public transit schedules, checking the current time, and comparing online product prices.
These tool-enhanced responses are well aligned with user-expected service types, enabling more informed decisions, such as whether to speed up when approaching a subway station, whether to purchase an item, and how to manage calorie intake and corresponding exercise, without manually checking their phone.
In contrast, baselines such as ContextLLM lack the ability to map sensory context to tool-calling decisions, leading to less useful proactive assistance.

\subsection{Effectiveness of System Module}
\label{sec:Effectiveness of System Module}

\subsubsection{Effectiveness of Proactive-Oriented Context Extraction}
This subsection evaluates whether the proactive-oriented context extraction module can efficiently and accurately derive contexts needed for subsequent reasoning.

\begin{figure}
\begin{minipage}[b]{0.49\columnwidth}
     \centering
\includegraphics[width=1\linewidth]{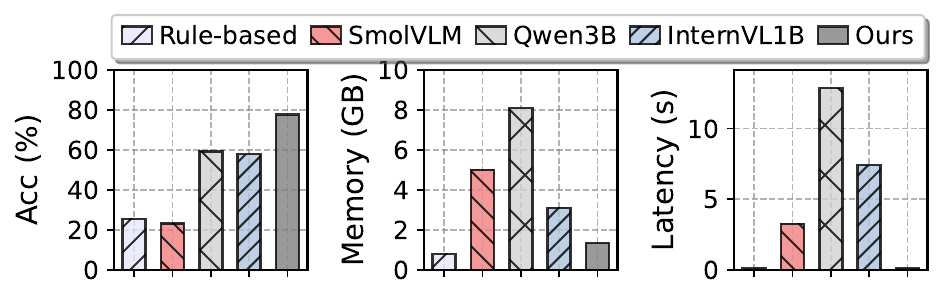}
\vspace{-2em}
  \caption{Performance of coarse-grained context extraction for context-aware persona retrieval.}
\label{fig:coarse_context_comparison}
\end{minipage}
\hfill
\begin{minipage}[b]{0.485\columnwidth} 
     \centering
\includegraphics[width=1\textwidth]{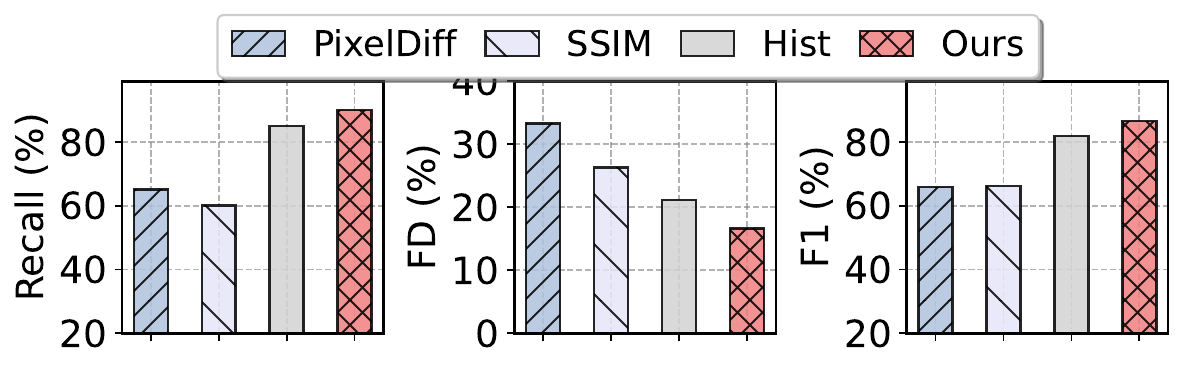}
\vspace{-2em}
  \caption{Comparison of different approaches for user focus verification.}
  \label{fig:results_focus}
\end{minipage}

\vspace{-1.em}
\end{figure}

\noindent\textbf{Coarse-grained Context Extraction.}
We first evaluate the coarse-grained vision context extraction in \workname. We compare our approach with one rule-based baseline and three VLM-based baselines. The rule-based baseline uses the same detection model as ours with predefined rules for scene inference, while the VLM baselines directly infer scenes from visual inputs. Specifically, we use SmolVLM~\cite{marafioti2024smolvlm}, Qwen2.5-VL-3B~\cite{Qwen2.5-VL}, and InternVL3-1B~\cite{zhu2025internvl3}. Figure~\ref{fig:coarse_context_comparison} demonstrates that \workname~achieves 18.3\% higher accuracy while using 6.0× less memory than Qwen2.5-VL-3B. Moreover, \workname~attains a latency of 118.4 ms for coarse-grained vision context extraction, which is substantially lower than most baselines. These results show that \workname~can efficiently and accurately extract visual contexts for subsequent persona retrieval.

\begin{figure}
\begin{minipage}[b]{0.32\columnwidth}
 \centering
\includegraphics[width=1\textwidth]
{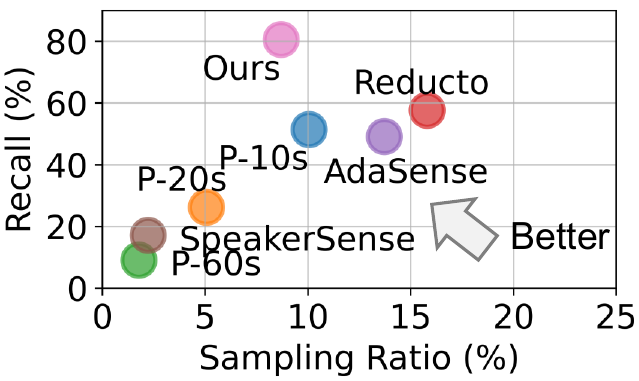}
\vspace{-2.5em}
  \caption{Performance of on-demand tiered perception in \workname.}
\label{fig:overall_efficiency_tolerance3}
\end{minipage}
\hspace{0.01\columnwidth} 
\begin{minipage}[b]{0.64\columnwidth}
     \centering
\includegraphics[width=1\linewidth]
{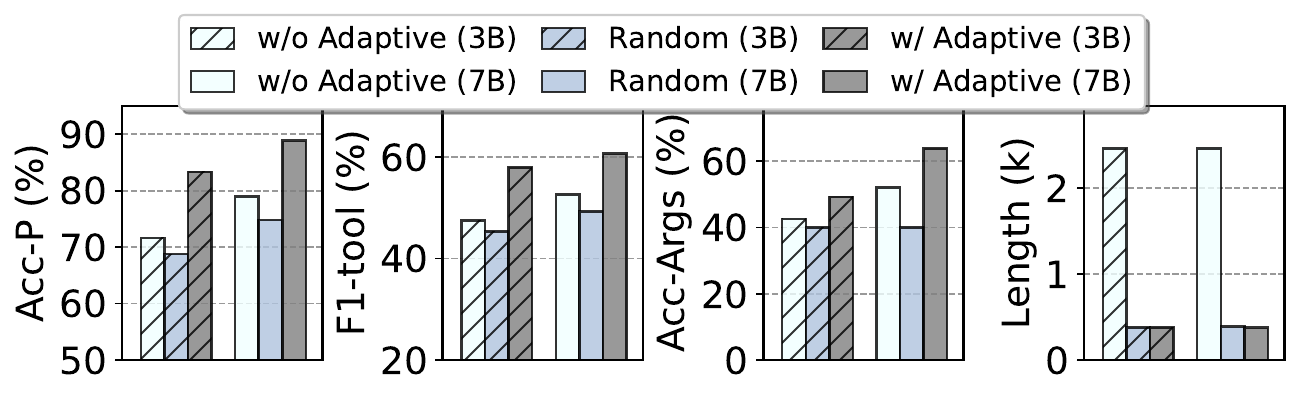}
\vspace{-2.5em}
  \caption{Performance of adaptive persona retrieval. Length denotes persona input length. Ada and Rand indicate adaptive retrieval and random selection.}
\label{fig:adaptive_retrieval_comparison}
\end{minipage}
\end{figure}

\noindent\textbf{Adaptive Persona Retrieval.}
We then evaluate the adaptive persona retrieval of \workname~on CAB-Lite.
\cmt{We compare its performance with using all personas from CAB-Lite and with random subsets of the same size.}
Figure~\ref{fig:adaptive_retrieval_comparison} shows that adaptive retrieval achieves up to 13.9\% higher \textit{Acc-P}, 4.4\% higher \textit{F1-tool}, and 3.8\% higher \textit{Acc-Args}.
It can also reduce input length by 6.5x, substantially lowering the overhead of agent reasoning.

\subsubsection{Effectiveness of On-Demand Tiered Perception}
\begin{figure}
    \centering
    \begin{subfigure}{0.335\columnwidth}
        \centering
        \includegraphics[width=1\linewidth]{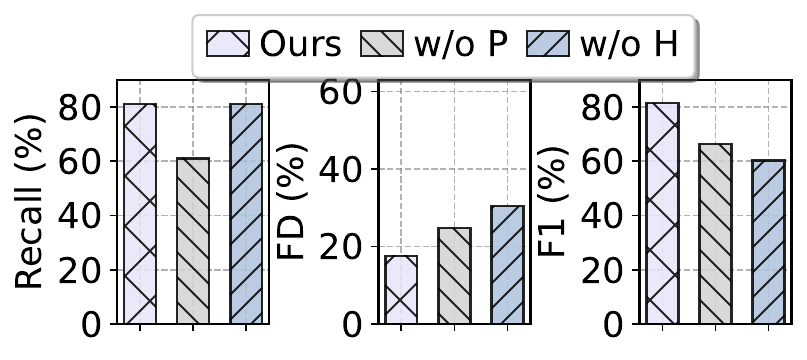}
\vspace{-2em}
  \caption{Impact of different contexts. }
\label{fig:ablation_overall_P_H}
    \end{subfigure}
    \begin{subfigure}{0.335\columnwidth}  
        \centering 
        \includegraphics[width=1\linewidth]{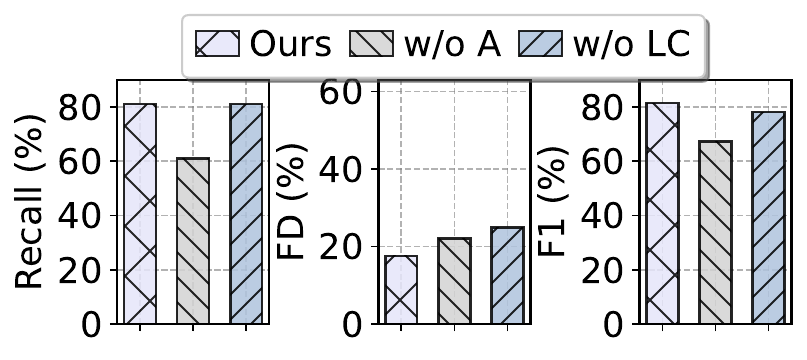}
\vspace{-2em}
  \caption{On-demand tiered perception. }
\label{fig:ablation_overall_A_LC}
    \end{subfigure}
    \begin{subfigure}{0.31\columnwidth}
        \centering
\includegraphics[width=1\textwidth]{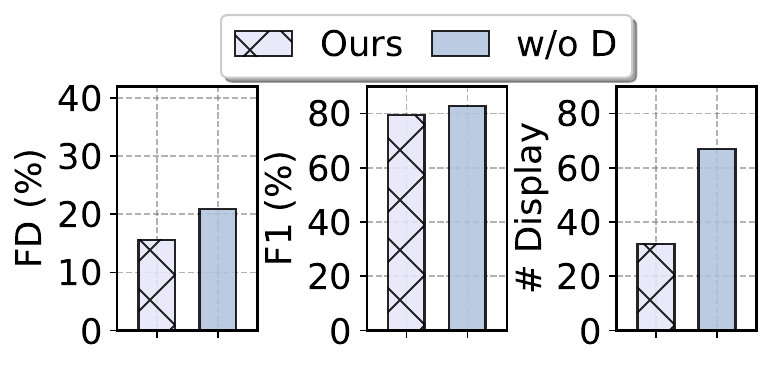}
\vspace{-2em}
  \caption{Proactive delivery control.}
  \label{fig:ablation_display}
    \end{subfigure}
    \vspace{-1em}
    \caption{Ablation study. ``w/o P'' and ``w/o H'' denote without persona and historical contexts, respectively. ``w/o A'' and ``w/o LC'' denote removing agent-in-the-loop adaptive perception and low-cost contextual cue triggering, respectively. ``w/o D'' denotes removing proactive delivery control.}
\label{fig:ablation_study}
      \vspace{-1.em}
\end{figure}

We evaluate the effectiveness of on-demand tiered perception in \workname~on the real-world dataset.
Specifically, we compare it with several strong baselines, including periodic vision sampling at fixed intervals of $x$ seconds (denoted as ``P-x''), Reducto~\cite{li2020reducto}, AdaSense~\cite{neseem2020adasense}, and SpeakerSense~\cite{lu2011speakersense}.
AdaSense and SpeakerSense trigger vision sampling using IMU motion states and detected speech, respectively.
Figure~\ref{fig:overall_efficiency_tolerance3} demonstrates that \workname~achieves the best trade-off between \textit{sampling ratio} and \textit{sampling recall}, validating the effectiveness of our design.
We also conduct an ablation study by removing low-cost contextual cue triggering (denoted as ``w/o LC'') and agent-in-the-loop adaptive perception (denoted as ``w/o A''). Figure~\ref{fig:ablation_overall_A_LC} demonstrates that removing these two designs reduces \textit{F1} by 3.1\% and 21.2\%, respectively, validating that both components contribute to the on-demand tiered perception.

\begin{figure}
  \centering
\includegraphics[width=1\linewidth]{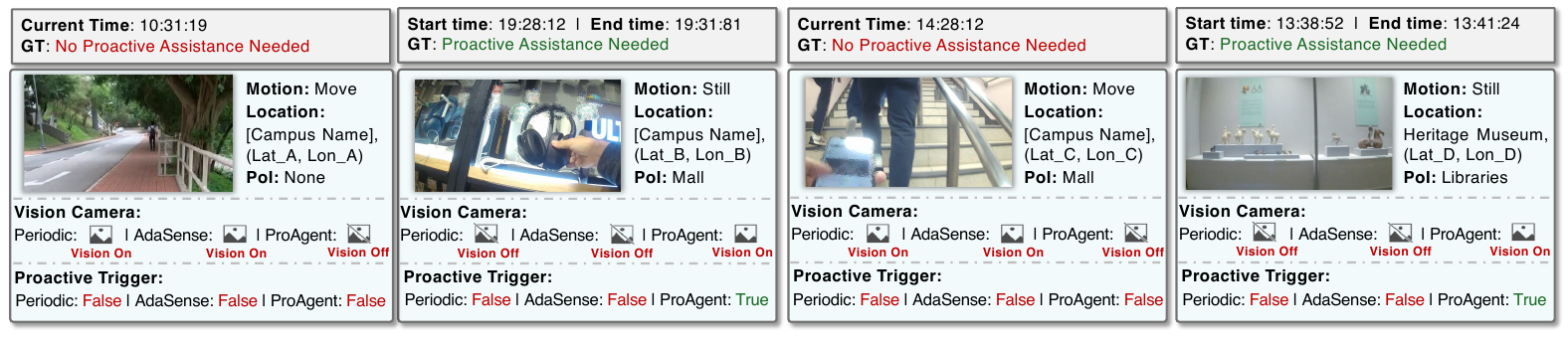}
\vspace{-2.5em}
  \caption{Examples of adaptive sampling approaches on real-world data. The top shows the ground truth (GT) proactive need at the current moment. \textit{Periodic} denotes periodic vision sampling. \textit{Vision camera} indicates whether visual perception is activated, while \textit{proactive trigger} denotes whether proactive assistance is triggered at that moment.}
  \vspace{-1.em}
\label{fig:sensor_trigger_time}
\end{figure}

We further provide examples of vision sampling decisions made by different approaches on real-world data. 
Figure~\ref{fig:sensor_trigger_time} illustrates that periodic sampling and AdaSense may capture redundant frames when the user does not need proactive assistance, such as street walking without notable events, yet miss critical moments that require assistance, such as shopping or museum visits. 
In contrast, \workname~uses agent-in-the-loop adaptive perception to suppress unnecessary visual sampling in low-value contexts and activate high-frequency perception when assistance is likely needed, improving both system efficiency and proactive timing accuracy.

\subsubsection{Effectiveness of Context-Aware Proactive Reasoner}
\label{sec:Effectiveness of Context-Aware Proactive Reasoner}
This subsection evaluates the context-aware proactive reasoner from three perspectives, including the impact of input contexts, the effectiveness of proactive delivery control, and the impact of context-aware CoT distillation.

\noindent\textbf{Impact of Contexts}.
First, we evaluate the impact of different contexts on \workname's proactive reasoning. Figure~\ref{fig:ablation_overall_P_H} illustrates the results on the real-world dataset when persona contexts (``w/o P'') and historical contexts (``w/o H'') are removed. 
Removing these contexts reduces \textit{F1} by 15.1\% and 21.2\%, respectively, validating the effectiveness of both contexts in proactive reasoning.
Personas are particularly helpful for improving \textit{Recall}, as some user-specific needs are difficult to infer without explicit preference cues, especially when the system is designed to avoid overly intrusive assistance. In contrast, historical context mainly helps reduce \textit{FD} by enabling the system to identify repeated or redundant assistance based on recent interactions.

\noindent\textbf{Impact of Proactive Delivery Control}.
Figure~\ref{fig:results_focus} shows that the user focus verification module in \workname~outperforms baselines based on PixelDiff, SSIM, and Hist.
Figure~\ref{fig:ablation_display} illustrates the results on the real-world dataset when proactive delivery control is removed (denoted as ``w/o D''). The results show that removing this module increases \textit{FD} by 5.3\% and nearly doubles the number of displayed reminders, while \textit{F1} remains almost unchanged.
This indicates that proactive delivery control effectively mitigates intrusiveness and excessive reminders.

\noindent\textbf{Impact of Thought Traces}.
Finally, we evaluate the impact of removing thought traces during SFT on the VLM agent's performance (denoted as ``w/o CoT'').
Figure~\ref{fig:ablation_vlm_combined} illustrates that \workname~achieves 3.6\% higher \textit{Acc-P}, 6.0\% higher \textit{F1-tool}, and 10.1\% higher \textit{Acc-Args}.
This improvement suggests that prompting the VLM reasoner to first generate thought traces from visual inputs helps it better interpret the current situation and infer user intent, leading to more accurate proactive predictions and tool use.

\subsubsection{Impact of Base VLMs.}
Next, we evaluate the performance of \workname~using different VLMs as the base model.
Figure~\ref{fig:base_vlm} illustrates that scaling up the base VLM consistently improves proactive reasoning performance. Notably, larger gains are observed in \textit{F1-tool} and \textit{Acc-Args} than in \textit{Acc-P}. For example, scaling the model from 7B to 32B improves \textit{F1-tool} by 3.1\%, while \textit{Acc-P} increases by only 1.1\%. 
This suggests that tool calling is more complex than the discriminative task of proactive need prediction, particularly for smaller VLMs.

\subsubsection{Impact of Hyperparameters.}
\begin{figure}
\begin{minipage}[b]{0.65\columnwidth}
 \centering
\includegraphics[width=1.0\columnwidth]{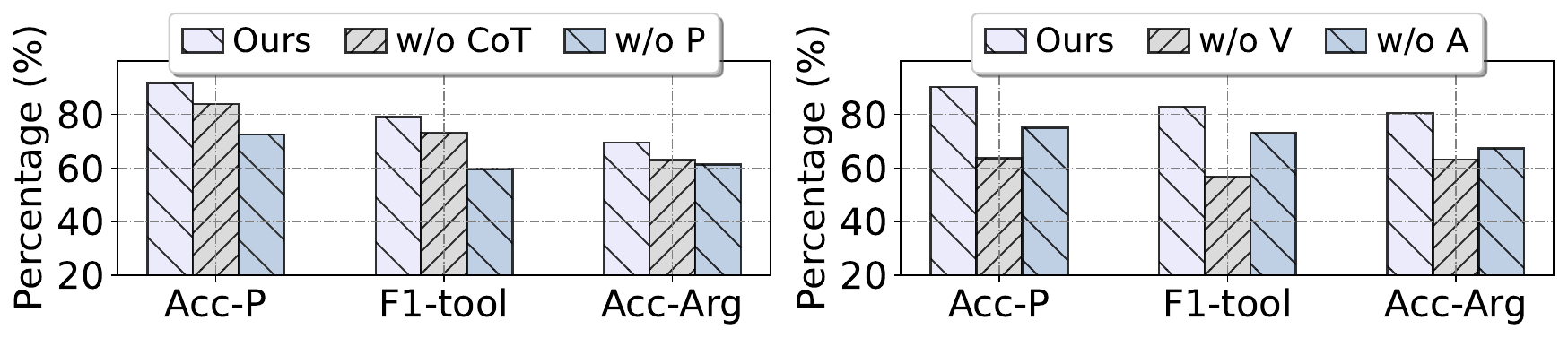}
    \vspace{-2.2em} 
    \caption{Ablation study on the CAB-Lite dataset. ``w/o CoT'' and ``w/o P'' denotes removing CoT reasoning and personas, respectively. ``w/o V'' and ``w/o A'' denotes removing vision and audio, respectively.} \label{fig:ablation_vlm_combined}
\end{minipage}
\begin{minipage}[b]{0.33\columnwidth}
     \centering
\includegraphics[width=0.88\columnwidth]{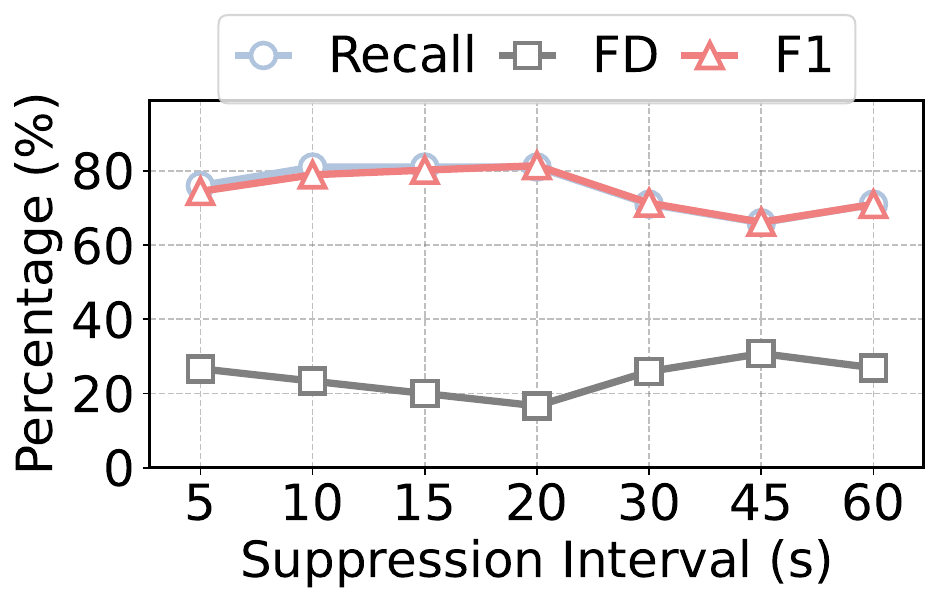}
    \vspace{-1em} 
    \caption{Impact of $\tau$} 
\label{fig:cue_cooldown_combined_lines}
\end{minipage}
\vspace{-1.em}
\end{figure}

\begin{figure}
\begin{minipage}[b]{0.49\columnwidth} 
     \centering
\includegraphics[width=1\textwidth]{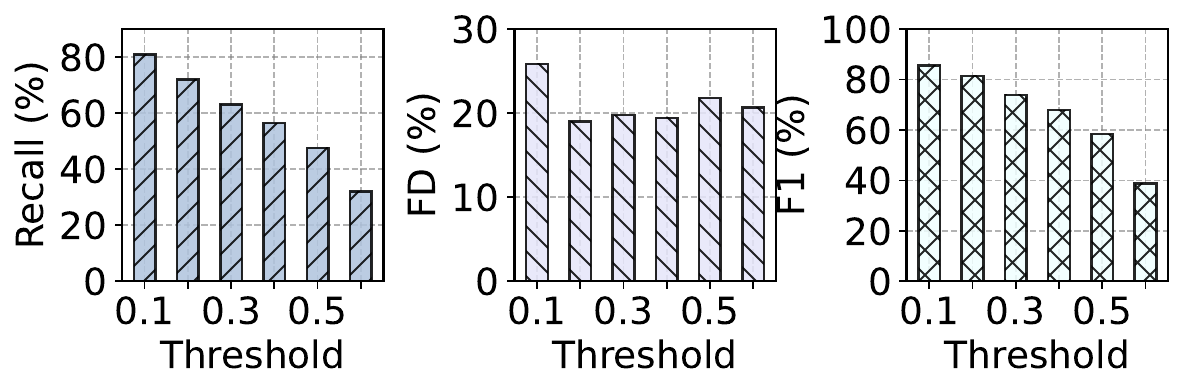}
\vspace{-2em}
  \caption{Impact of $\tau_f$.}
  \label{fig:hyper-parameter_pair_orb_match_ratio_all_metrics}
\end{minipage}
\hfill
  \begin{minipage}[b]{0.49\columnwidth}
     \centering
\includegraphics[width=1\textwidth]{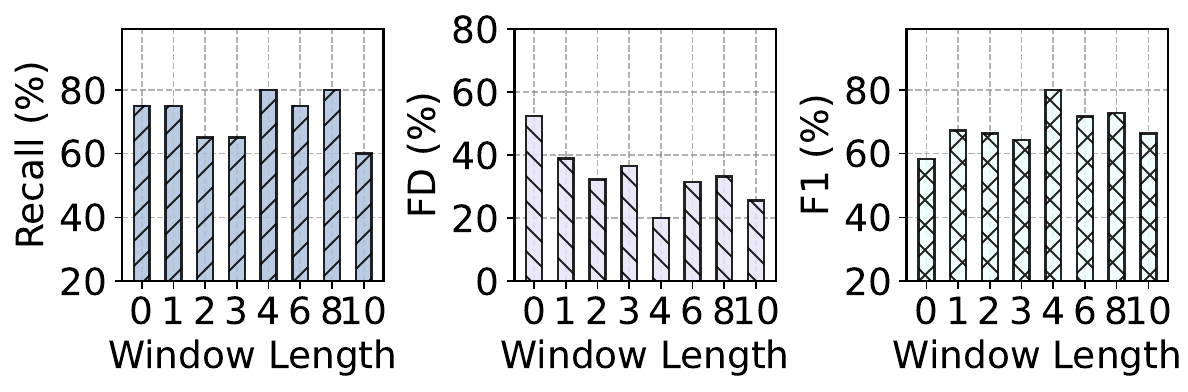}
\vspace{-2.em}
  \caption{Impact of $L$.}
\label{fig:history_sweep}
\end{minipage}
\vspace{-1.5em}
\end{figure}

We also evaluate the impact of hyperparameters in \workname. Figure~\ref{fig:hyper-parameter_pair_orb_match_ratio_all_metrics} shows the impact of $\tau_f$ in proactive delivery control. A smaller $\tau_f$ improves \textit{recall} but also increases \textit{FD}, and $\tau_f = 0.2$ achieves the best trade-off.
Figure~\ref{fig:history_sweep} shows the impact of history window length. We observe that $L=4$ achieves the lowest \textit{FD} while also achieving the highest \textit{F1}.
Figure~\ref{fig:cue_cooldown_combined_lines} shows the effect of the suppression interval $\tau$, where $\tau = 20 \text{s}$ achieves the lowest \textit{FD} and the highest \textit{F1}.
Therefore, we set $\tau_f = 0.2$, $L = 4$, and $\tau=20 s$ in \workname.

\subsection{System Overhead}
\label{sec:system_overhead}
We evaluate the overhead of \workname~across submodule inference time and communication latency on multiple hardware platforms and in outdoor scenarios.
Figure~\ref{fig:latency} shows that \workname~achieves reasoning times of 0.52 s on the A6000 and 3.89 s on the Jetson Orin.
Notably, the Time-To-First-Token (TTFT) on both Jetson Orin and laptop (RTX 1660 Ti GPU) is under 300ms.
However, the average participant-annotated acceptable window for proactive assistance is only 2.8 minutes. These results suggest that \workname~can deliver timely proactive assistance when deployed on edge servers.
The personas occupy 86.1KB of storage.
Additionally, coarse-grained context extraction (CE) and temporal constraints (TC) incur 118.5ms and 58.6ms on Jetson Orin, respectively.
\workname~achieves average communication latencies of 321.2ms, 434.1ms, and 726.5ms in office, store, and subway scenarios, validating real-world usability. 
In addition, \workname~increases hourly battery consumption by only 4.2\% on the Google Pixel 7 and 12.1\% on the RayNeo X3 Pro compared with the idle condition, indicating it does not impose significantly more power overhead than regular use.

\subsection{Understanding \workname's Proactive Performance}
In this section, we provide a detailed analysis of \workname's performance, including error analysis of false detections and the impact of integrating the personas inferred from human feedback.

\subsubsection{Proactive Triggering Behavior}
\label{sec:Proactive Triggering Behavior}
\begin{figure}
  \centering
\includegraphics[width=1\linewidth]{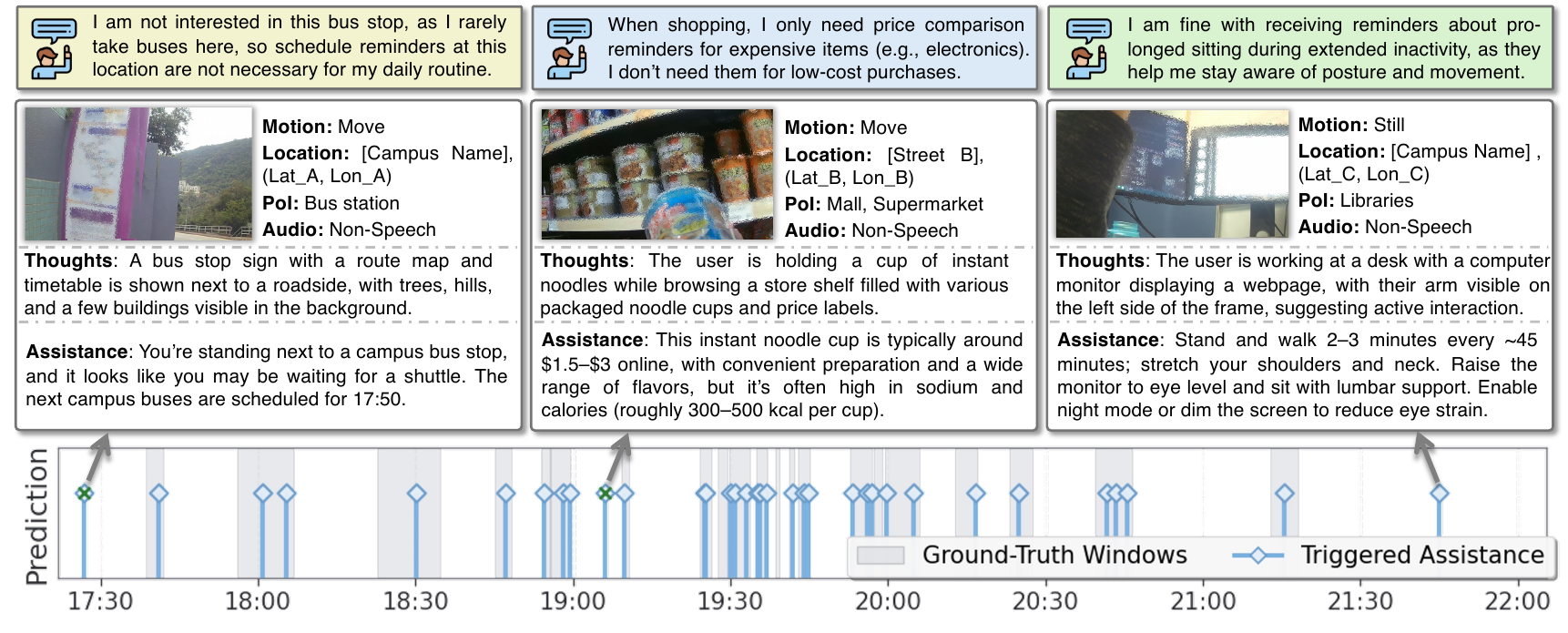}
\vspace{-2.5em}
  \caption{Examples of false detections by \workname~on real-world data. Each example includes the sensory context, the corresponding response, and a brief explanation shown at the top.
  Green $\times$ marks denote false detections that can be mitigated through the personas inferred from human feedback.}
  \vspace{-1.em}
\label{fig:end2end_response_false_alarm_v1}
\end{figure}

In this section, we analyze the underlying causes of false detections in \workname~and discuss potential mitigation strategies.
Figure~\ref{fig:end2end_response_false_alarm_v1} illustrates several false detection examples from real-world data, including the sensory context, \workname's assistance, and brief explanations of these cases. 
Results show that \workname~may provide proactive assistance (e.g., price comparisons) when users browse products. However, for this particular user, such assistance is only needed for high-value items (e.g., electronics and smartphones) rather than low-cost everyday items.
It may also suggest bus schedules at stops that users rarely use or are not interested in. 
These cases indicate that user-specific preferences can lead to false detections in the default \workname, while results in Figure~\ref{fig:end2end_response_false_alarm_v1} show that they can be mitigated through the human feedback mechanism (see \S~\ref{sec:Impact of Human Feedback}).
In addition, low-quality sensory context, such as motion blur or occlusion due to movement, can lead to errors in VLM reasoning, resulting in both false and missed detections.
Some false detections may arise from the subjectivity of human annotations. 
During annotation, users may not explicitly mark all moments where proactive assistance could be beneficial, leading to unannotated but still acceptable opportunities.
However, when these proactive responses are presented to users, they may still consider them reasonable and useful.
For example, participants may not mark a sedentary reminder as necessary during annotation, but later consider it acceptable after seeing the generated response. Therefore, we conduct a user study in \S~\ref{sec:user_study} for a more comprehensive evaluation of \workname's proactive performance.

\begin{figure}
    \centering
     \begin{subfigure}{0.49\columnwidth}  
    \centering \includegraphics[width=1.0\columnwidth]{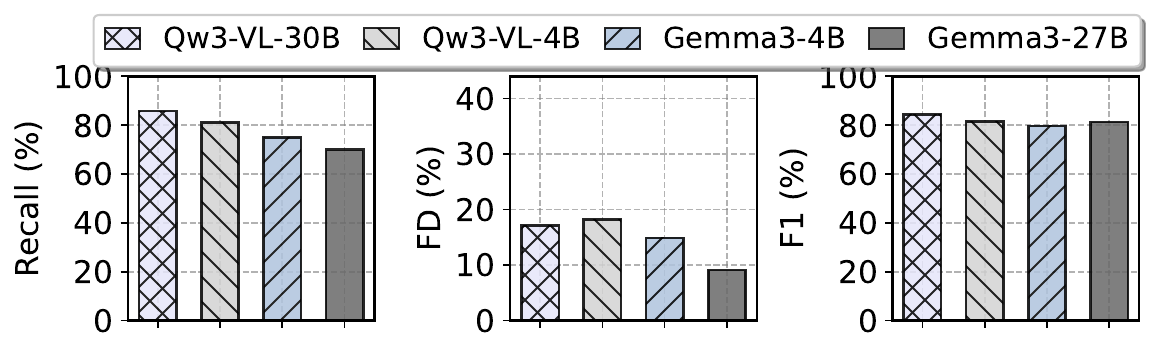}
            \vspace{-2.em} 
    \caption{Real-world dataset.} 
    \label{fig:base_vlm_real_world_all}
    \end{subfigure}
    \hspace{0.005\columnwidth} 
    \begin{subfigure}{0.49\columnwidth}  
    \centering \includegraphics[width=1.0\columnwidth]{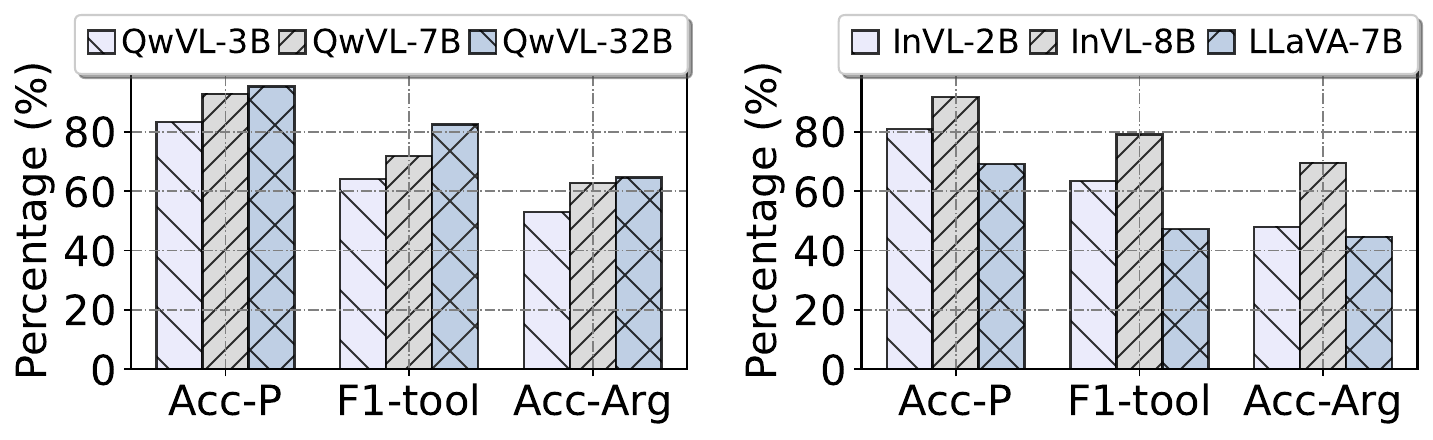}
            \vspace{-2.em} 
    \caption{CAB-Lite dataset.} 
    \label{fig:base_vlm_cablite}
    \end{subfigure}
   \vspace{-1em} 
    \caption{Comparison of different base VLMs used in \workname. Qw3VL is Qwen3-VL. InVL is InternVL.}
\label{fig:base_vlm}
\vspace{-1.em} 
\end{figure}

\begin{figure}
\begin{minipage}[b]{0.65\columnwidth} 
     \centering
\includegraphics[width=1\textwidth]{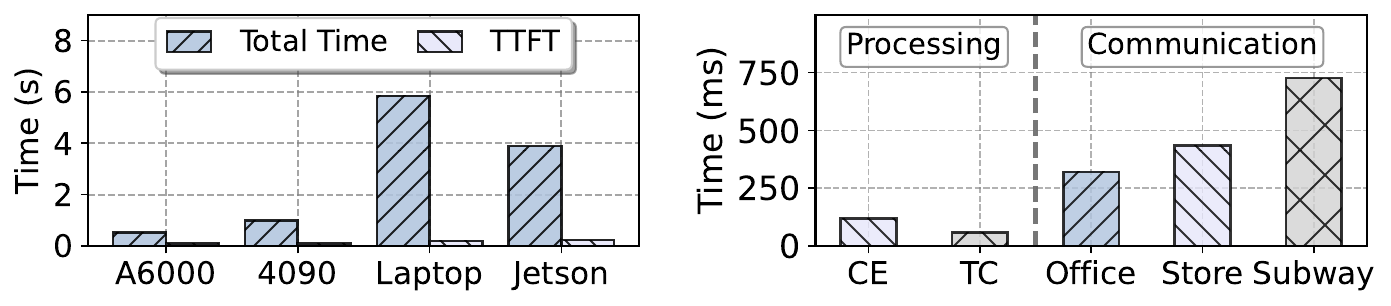}
\vspace{-2.2em}
  \caption{System latency.}
  \label{fig:latency}
\end{minipage}
 \hspace{0.02\columnwidth} 
\begin{minipage}[b]{0.28\columnwidth}
     \centering
\includegraphics[width=0.97\columnwidth]{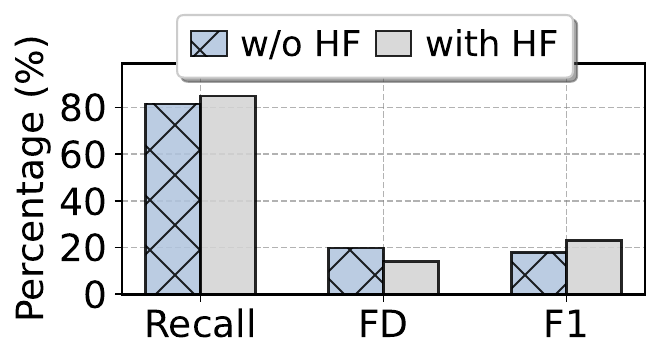}
\vspace{-1em}
\caption{{Human feedback impact}.} 
\label{fig:persona_bar}
\end{minipage}
\vspace{-1em}
\end{figure}

\subsubsection{Impact of Human Feedback}
\label{sec:Impact of Human Feedback}
Next, we evaluate the effectiveness of using personas inferred from user feedback for proactive assistance.
Figure~\ref{fig:base_vlm} compares \workname~with and without feedback-derived personas. The results show that such personas help correct some false detections by reducing unnecessary proactive assistance. Figure~\ref{fig:persona_bar} further shows that feedback-derived personas reduce \textit{FD} by 5.8\% and improve \textit{F1} by 5.1\%.
This improvement comes from \workname's ability to derive context-aware personas by integrating users' natural verbal feedback with the corresponding sensory context.
For example, if a user verbally indicates that they do not care about school bus schedule information at a specific bus stop, \workname~combines this feedback with contextual cues such as the current GPS location (Lat A, Lon A) and employ an LLM to infer a context-aware persona, such as ``\textit{At bus stops near GPS location (Lat A, Lon A), the user does not want bus schedule notifications when they are just passing by.}''
When the user later encounters a similar context, this persona helps \workname~suppress unnecessary proactive assistance.

\subsection{User Study}
\label{sec:user_study}

We conducted a user study with the same 20 participants from our real-world data collection to evaluate the proactive assistance provided by \workname~and baselines. Participants included 12 males and 8 females with an average age of 24.3 years, with further details described in \S~\ref{sec:dataset}.
Participants evaluated the proactive assistance generated by different approaches based on real-world sensory contexts captured by our prototype.
Each participant then rated the assistance using a 5-point Likert scale across the following six questions.

\begin{itemize}[leftmargin=*]
    \item \textbf{\textit{Q1}}: Have you previously used or interacted with any proactive personal assistant systems?
    
    \item \textbf{\textit{Q2}}: Was the timing of the system's proactive assistance appropriate when it was delivered?
    
    \item \textbf{\textit{Q3}}: Was the content of the proactive assistance provided by the system useful for the situation?
    
    \item \textbf{\textit{Q4}}: Did the system feel intrusive or unnecessarily interruptive when providing assistance?
    
    \item \textbf{\textit{Q5}}: Overall, how satisfied were you with the experience of using this system?
    
    \item \textbf{\textit{Q6}}: Would you be willing to use such a system in your daily life?

\end{itemize}

\begin{figure}
    \centering
    \begin{subfigure}{0.32\columnwidth}
        \centering
        \includegraphics[width=0.85\textwidth]{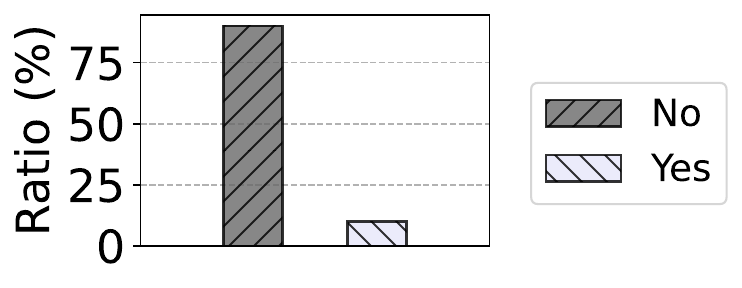}
        \vspace{-0.9em}
        \caption{Have you previously used or interacted with any proactive personal assistant systems?}  \label{fig:user_study_Q1}
    \end{subfigure}
    \hfill
    \begin{subfigure}{0.32\columnwidth}  
        \centering 
        \includegraphics[width=1\textwidth]{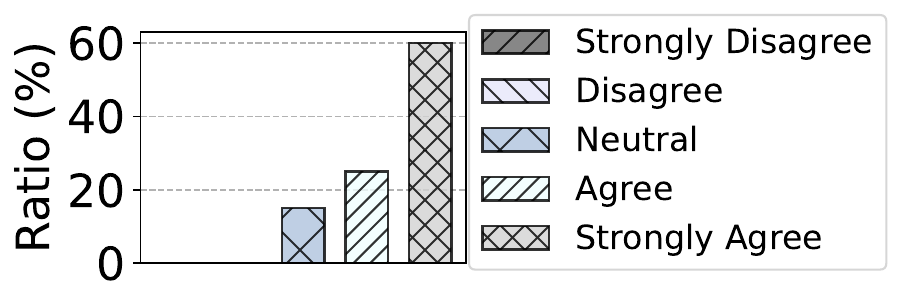}
        \vspace{-2.0em}
        \caption{Was the timing of the system's proactive assistance appropriate when it was delivered?}    
        \label{fig:user_study_Q2}
    \end{subfigure}
    \hfill
    \begin{subfigure}{0.32\columnwidth}  
        \centering 
        \includegraphics[width=1\textwidth]{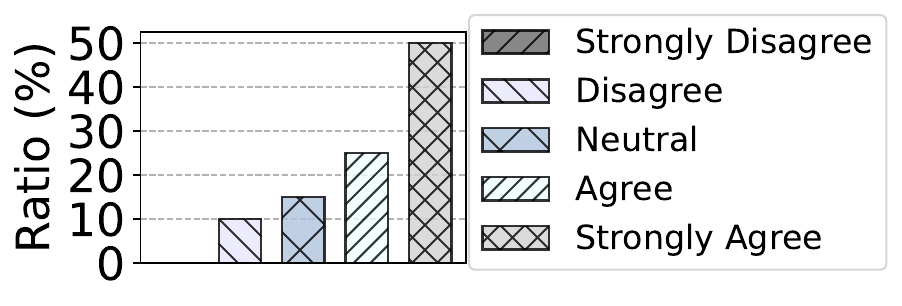}
        \vspace{-2.0em}
        \caption{Was the content of the proactive assistance provided by the system useful for the situation?}    
        \label{fig:user_study_Q3}
    \end{subfigure}
    \vspace{-1.0em}
    \vskip\baselineskip
    \begin{subfigure}{0.32\columnwidth}   
        \centering 
    \includegraphics[width=1\textwidth]{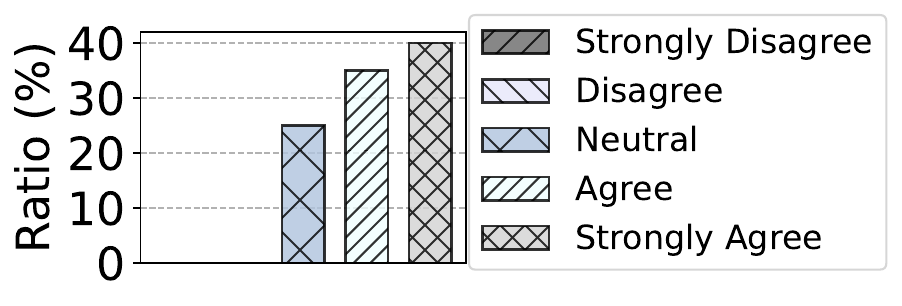}
        \vspace{-2.0em}
        \caption{Did the system feel intrusive when providing assistance?}    \label{fig:user_study_Q4}
    \end{subfigure}
    \hfill
    \begin{subfigure}{0.32\columnwidth}   
        \centering 
        \includegraphics[width=1\textwidth]{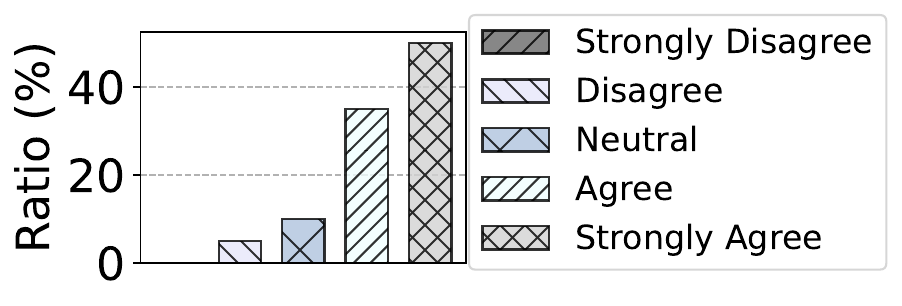}
        \vspace{-2.0em}
        \caption{Overall, how satisfied were you with the experience of using this system?}    \label{fig:user_study_Q5}
    \end{subfigure}
    \hfill
    \begin{subfigure}{0.32\columnwidth}  
        \centering 
    \includegraphics[width=1\textwidth]{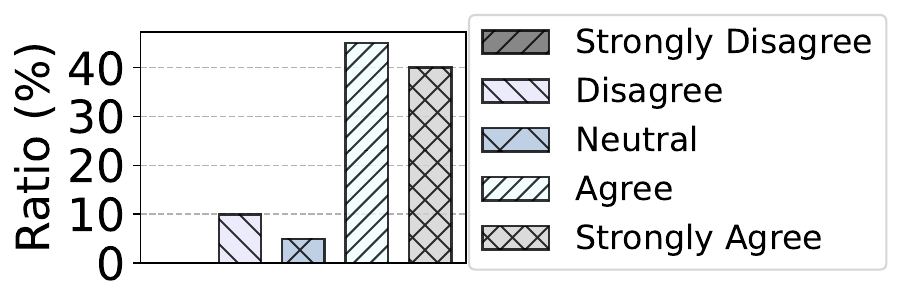}
        \vspace{-2.0em}
        \caption{Would you be willing to use such a system in your daily life?}    
        \label{fig:user_study_Q6}
    \end{subfigure}
     \vspace{-.8em}
    \caption{Participants' feedback on the proactive assistance performance of \workname.}
\label{fig:user_study_questionaire}
      \vspace{-.5em}
\end{figure}

Figure~\ref{fig:user_study_questionaire} illustrates participants' ratings of \workname.
The results demonstrate that 90\% of participants had no prior experience with proactive personal assistant systems, suggesting that proactive assistance remains unfamiliar to most users in daily life.
More than 85\% agreed that \workname~delivered assistance at appropriate moments, while the remaining participants were neutral.
For content usefulness, 75\% agreed that the assistance was useful for the current situation, and only 10\% considered it not sufficiently useful.
In terms of intrusiveness, 75\% reported that the assistance was not intrusive or disruptive, while the remaining 25\% were neutral.
Participants also showed positive overall attitudes toward using \workname. Overall, 85\% of participants reported positive attitudes toward the experience and their willingness to use \workname~in daily life. In contrast, 10\% reported neutral attitudes, and only 5\% expressed dissatisfaction. These results indicate that \workname~was generally well received and shows potential for daily-life proactive assistance.

Participants valued proactive assistance that provided actionable and informative support, rather than generic questions or broad suggestions.
For example, several participants appreciated dining-related suggestions, as they often struggled to decide what to eat. Others found shopping-related assistance useful when \workname~provided discount information or online price comparisons, especially for relatively expensive products. Transportation-related assistance, such as subway or shuttle bus arrival reminders, was also considered helpful. Some participants further suggested that crowd-level information could improve such assistance by helping them decide where to wait or which carriage to choose. These comments suggest that \workname~provides timely, concrete, and actionable assistance that helps reduce users’ decision-making effort in daily scenarios.

\begin{figure}
    \centering
    \begin{subfigure}{0.3\columnwidth}  
    \centering \includegraphics[width=1.0\columnwidth]{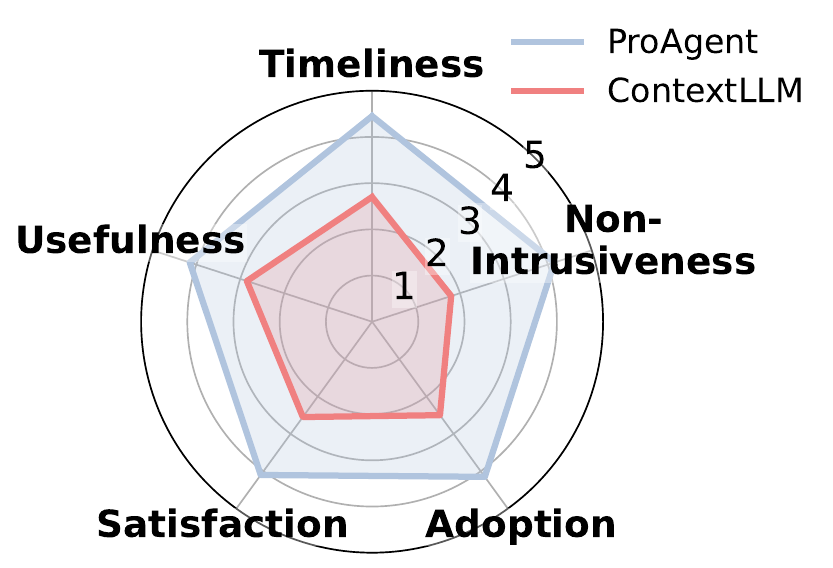}
            \vspace{-1.5em} 
    \caption{ContextLLM.} \label{fig:user_study_radar_contextllm}
    \end{subfigure}
    \begin{subfigure}{0.33\columnwidth}  
    \centering \includegraphics[width=1.0\columnwidth]{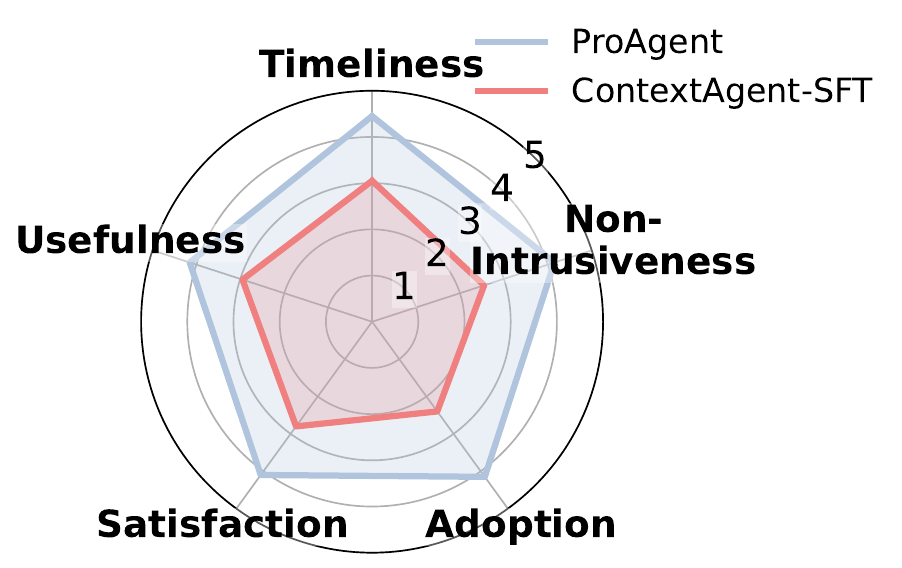}
            \vspace{-1.5em} 
    \caption{ContextAgent-SFT.} \label{fig:user_study_radar_ContextAgent}
    \end{subfigure}
    \begin{subfigure}{0.3\columnwidth}  
    \centering \includegraphics[width=1.0\columnwidth]{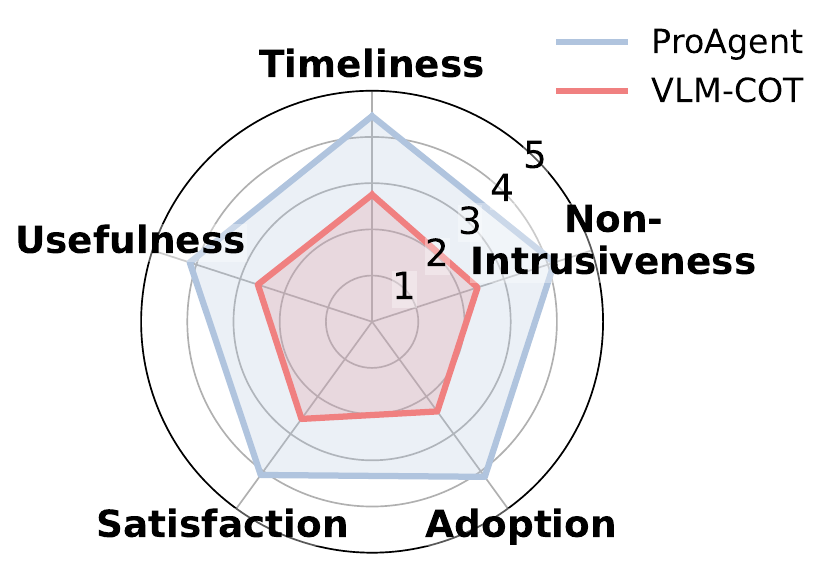}
            \vspace{-1.5em} 
    \caption{VLM-CoT.} \label{fig:user_study_radar_cot}
    \end{subfigure}
   \vspace{-1em} 
    \caption{User study. Comparison of human ratings for \workname~and baselines.}
\label{fig:user_study_radar}
\vspace{-1.7em} 
\end{figure}

Figure~\ref{fig:user_study_radar} demonstrates the comparison of human ratings between \workname~and baselines.
We select three representative approaches, including ContextLLM, ContextAgent-SFT, and VLM-CoT, as they cover different categories of designs and achieve higher performance in our quantitative evaluation.
The five dimensions in Figure~\ref{fig:user_study_radar} correspond to Q2–Q6 in the questionnaire.
Results show that \workname~receives consistently higher ratings than the baselines across all five dimensions.
Several participants perceived baseline approaches as more intrusive because they provided assistance too frequently or in less relevant situations. For example, some baselines triggered assistance for subway advertisements even when participants were focused on their phones and did not need such information, suggesting a limited ability to infer the user’s current intent.
Participants also found some baseline outputs less actionable, as they often asked whether the user needed help rather than directly providing useful information. In contrast, participants preferred assistance that offered concrete support, such as arrival times and discount information.

The feedback further reveals several design implications for daily-life proactive agents. First, proactive assistance should not be triggered simply by the presence of relevant objects or contexts. Instead, the system should consider intent-related cues that indicate a potential need for assistance. 
Several participants appreciated that \workname~did not frequently interrupt them while they were simply browsing, but provided shopping-related assistance when they showed interest in a product. 
This suggests that proactive systems should distinguish genuine user interest from incidental contextual cues. 
Second, assistance should be concrete and actionable. Participants preferred concrete information, such as arrival times, discount information, and price comparisons, over generic questions asking whether they needed help, which were often perceived as unnecessary interruptions.
Third, the value of proactive assistance varies with individual preferences. Some participants valued price-comparison assistance for expensive products but found it unnecessary for low-cost everyday goods, while others felt such assistance was promotional or preferred to check prices themselves. 
Some participants preferred fewer interruptions, whereas others expressed interest in follow-up interactions.
These differences highlight the need to update personas through user feedback, enabling the system to capture user-specific preferences and decide what assistance to provide and when to provide it. Participants also mentioned presentation issues, such as overly long messages or insufficient display time, suggesting that proactive assistance should be concise and easy to consume in daily use.

%% file: secs/6_discussion.tex
\section{Discussion}
\noindent\textbf{Scalability of \workname.}
\workname~currently adopts an API-based implementation for agent tool calling. 
Recent advances in reasoning-enhanced perception (e.g., reinforcement learning~\cite{yu2025perception}), Model Context Protocol (MCP)~\cite{hou2025model}, and efficient LLM inference techniques~\cite{yang2026efficient,leviathan2023fast,gu2023mamba,yang2023edgefm} are orthogonal to this work. These techniques can be integrated into \workname~to further improve its reasoning capability, tool extensibility, and system efficiency.

\noindent\textbf{Privacy Concerns.}
\workname~can be run on personal devices or edge servers like laptops without cloud access, ensuring user data remains local and preserving privacy. 
\workname~can also use hardware-based cues, such as visible indicators, flashing lights, or audio alerts, to notify bystanders when environmental sensing is active.

\noindent\textbf{Power Constraints.}
Current commercial smartglasses still have limited battery capacity, which makes long-duration in-the-wild deployment challenging. Although \workname's on-demand tiered perception reduces the energy cost of visual sensing by activating high-rate vision only when needed, relying solely on the built-in battery of smartglasses may still be insufficient for extended daily use. To mitigate this limitation, our prototype uses external power capsules to support longer-duration deployment in real-world scenarios. We expect future advances in low-power chips, battery capacity, and AR glasses hardware to further alleviate this constraint.